\documentclass[11pt,letterpaper]{article}
\usepackage[T1]{fontenc}
\usepackage[utf8]{inputenc}
\usepackage{lmodern}
\DeclareUnicodeCharacter{00B2}{\ensuremath{^2}}
\DeclareUnicodeCharacter{039B}{\ensuremath{\Lambda}}
\DeclareUnicodeCharacter{03B3}{\ensuremath{\gamma}}
\DeclareUnicodeCharacter{03BB}{\ensuremath{\lambda}}
\DeclareUnicodeCharacter{03C0}{\ensuremath{\pi}}
\DeclareUnicodeCharacter{03F1}{\ensuremath{\rho}}
\DeclareUnicodeCharacter{2013}{--}
\DeclareUnicodeCharacter{2019}{'}
\DeclareUnicodeCharacter{2212}{-}
\DeclareUnicodeCharacter{221E}{\ensuremath{\infty}}
\DeclareUnicodeCharacter{2264}{\ensuremath{\leq}}
\DeclareUnicodeCharacter{2265}{\ensuremath{\geq}}
\usepackage[margin=1in]{geometry}
\usepackage{amsmath,amssymb,amsthm}
\usepackage{microtype}
\usepackage{times}
\usepackage{booktabs,longtable,array,ragged2e,multirow,color,tabularx}
\usepackage{graphicx,float,placeins}
\usepackage{caption}
\usepackage{subcaption}
\usepackage{enumitem}
\usepackage{titletoc}
\usepackage[round,authoryear]{natbib}
\usepackage[hidelinks]{hyperref}
\hypersetup{colorlinks=true,citecolor=blue,linkcolor=blue,urlcolor=blue,
  pdftitle={The Gold Rush in AI4Math: Signatures of AI-assisted Proofs}}
\usepackage{url}
\title{\textbf{The Gold Rush in AI4Math, II: What are signatures of AI-assisted Proofs?}}
\author{Yuanchuan Guo\footnote{yguo1@g.harvard.edu} \and Jiashun Jin\footnote{jiashunmail@gmail.com} \and Zheng Tracy Ke\footnote{zke@fas.harvard.edu} \and Bingcheng Sui\footnote{bingchengsui@g.harvard.edu}}
\date{}
\begin{document}
\maketitle

\begin{abstract}
The growing use of AI in mathematics raises a question: can we detect AI involvement in proofs, separately from AI assistance with writing? Historical arXiv papers, recent papers disclosing substantive AI use, and manuscripts released by OpenAI provide a rich empirical basis for studying this question. However, several challenges remain: there is a lack of matched case-control pairs; disclosure of AI use can be incomplete or imprecise; and we do not yet have ready-to-use features specifically designed to capture signals of AI involvement in proofs.

To address these challenges, we construct five cohorts comprising 8,281 papers by combining existing manuscripts with those we produced using generative AI. These cohorts span different levels and forms of AI involvement in mathematics and include matched pairs of papers addressing the same research problems. We then introduce ProofTrace, a framework that represents each paper as a graph of immediate proof support (GIPS) and extracts interpretable features of mathematical organization. Using these cohorts and this framework, we identify statistically significant structural features (``signatures") that distinguish AI-generated from human-written proofs.

Our study yields three major discoveries. First, principal-component projections of our features reveal a qualitative gradient of AI involvement, where AI-reworded papers remain close to human papers, the two AI-generated cohorts depart in broadly shared directions, and papers disclosing substantive AI use occupy intermediate positions. The latter cohort also shifts toward the AI-rebuilt reference over time. Second, we identify recurring signatures of AI-generated proofs, such as a smaller proportion of externally cited results, greater reliance on intermediate conclusions developed within proofs rather than stated as lemmas, and altered graph connectivity; many of these signatures persist under controls for research field, manuscript length, or graph size. Finally, in a separate detection experiment, ProofTrace distinguishes human-written papers from those disclosing substantive AI use. Its performance remains stable under AI rewording, whereas the performance of several state-of-the-art text-based detectors declines sharply.
\end{abstract}

\section{Introduction}

There is growing interest in using artificial intelligence (AI) in mathematical research, from formulating problems to developing proofs. A recent study of approximately 33,000 arXiv submissions in mathematics reports a sharp increase in self-disclosed substantive AI use \citep{JinKeSui2026}. In September--October 2026, OpenAI released a collection of mathematical manuscripts produced by its internal model \citep{OpenAI2026MathRelease},  generating extensive discussion and debate within the academic community and on social media.

As these uses expand, a fundamental question emerges: can we distinguish AI involvement in proofs from AI assistance with writing? Answering this question could encourage accurate disclosure of AI use, help researchers understand and assess AI-generated proofs, and guide the evaluation and improvement of AI systems for mathematical research.

One challenge is the lack of suitable datasets. Available sources include papers posted on arXiv, some of which include AI disclosure statements, and manuscripts released by OpenAI. These papers span different levels of AI involvement, but differences in the research problems they address may confound comparisons. In addition, authors' disclosures can be incomplete or imprecise. The absence of a disclosure does not imply that a paper's mathematics was developed without AI. Even when AI assistance is disclosed, the disclosure generally does not specify the proportion of mathematical content to which AI contributed.

To address this challenge, we construct a dataset comprising 8,281 papers across five cohorts that span different forms and levels of AI involvement. The dataset includes two complementary paired subsets. One pairs human-written papers with AI-reworded versions intended to preserve their mathematical content; the other pairs human-written papers with AI-rebuilt manuscripts generated from briefs specifying the same research settings and target conclusions. These matched pairs reduce confounding by research topic and help distinguish AI involvement in proof development from AI assistance with writing.

Another challenge is the lack of ready-to-use features tailored to AI involvement in proofs. Existing approaches to AI-text detection \citep[e.g.,][]{gehrmann2019gltr,ippolito2020automatic, mitchell2023detectgpt} typically rely on linguistic features or representations extracted from pretrained language models. Linguistic features primarily characterize writing, while pretrained representations may combine signals from mathematical content and linguistic presentation. We therefore seek a feature space that directly characterizes mathematical reasoning and helps distinguish AI involvement in proof development from AI assistance with writing.

To address this challenge, we introduce ProofTrace, a feature extraction framework that represents each paper as a \emph{graph of immediate proof support (GIPS)}. Each node represents a stated result or a substantive intermediate conclusion, and a directed edge from $A$ to $B$ records that the proof of $B$ directly uses the conclusion of $A$. This representation captures which mathematical results appear in a paper and how they are connected through proof dependencies. We then extract six categories of interpretable features from the graph and its attributes, providing a feature bank for downstream statistical analysis.

Our statistical analyses using the constructed dataset and ProofTrace features yield three main findings. First, the ProofTrace feature space exhibits a qualitative gradient of AI involvement: AI-reworded papers remain close to the Human reference, while the two AI-generated cohorts depart in broadly similar directions. The Hybrid cohort occupies intermediate positions and shifts toward the AI-rebuilt reference over the study period. Second, paired comparisons between AI-rebuilt papers and their human-written counterparts reveal statistically significant differences in features such as the proportion of externally cited results, the prevalence of intermediate conclusions, and graph connectivity. Many of these differences persist after accounting for research field, manuscript length, or graph size. Third, in a detection experiment, a classifier trained on ProofTrace features maintains its performance under AI rewording that preserve the mathematics. It suggests that the classifier captures signals of AI involvement in proof development rather than in writing. 

Our study contributes to recent efforts to understand the growing use of AI across scientific disciplines \citep[e.g.,][]{liang2025quantifying,cheng2024ai,hao2026ai,he2026policies}. Much of this work focuses on AI-assisted writing, whereas we examine AI involvement in proof development, a form of assistance that directly shapes the intellectual content of research. Although our study centers on mathematics, our framework could be adapted to other fields that rely on mathematical reasoning, including theoretical physics, econometrics, probability, statistics, and machine learning.


\section{Five cohorts and paired study design}
\label{sec:cohort-design}

\begin{figure}[htb]
\centering
\includegraphics[width=.82\linewidth]{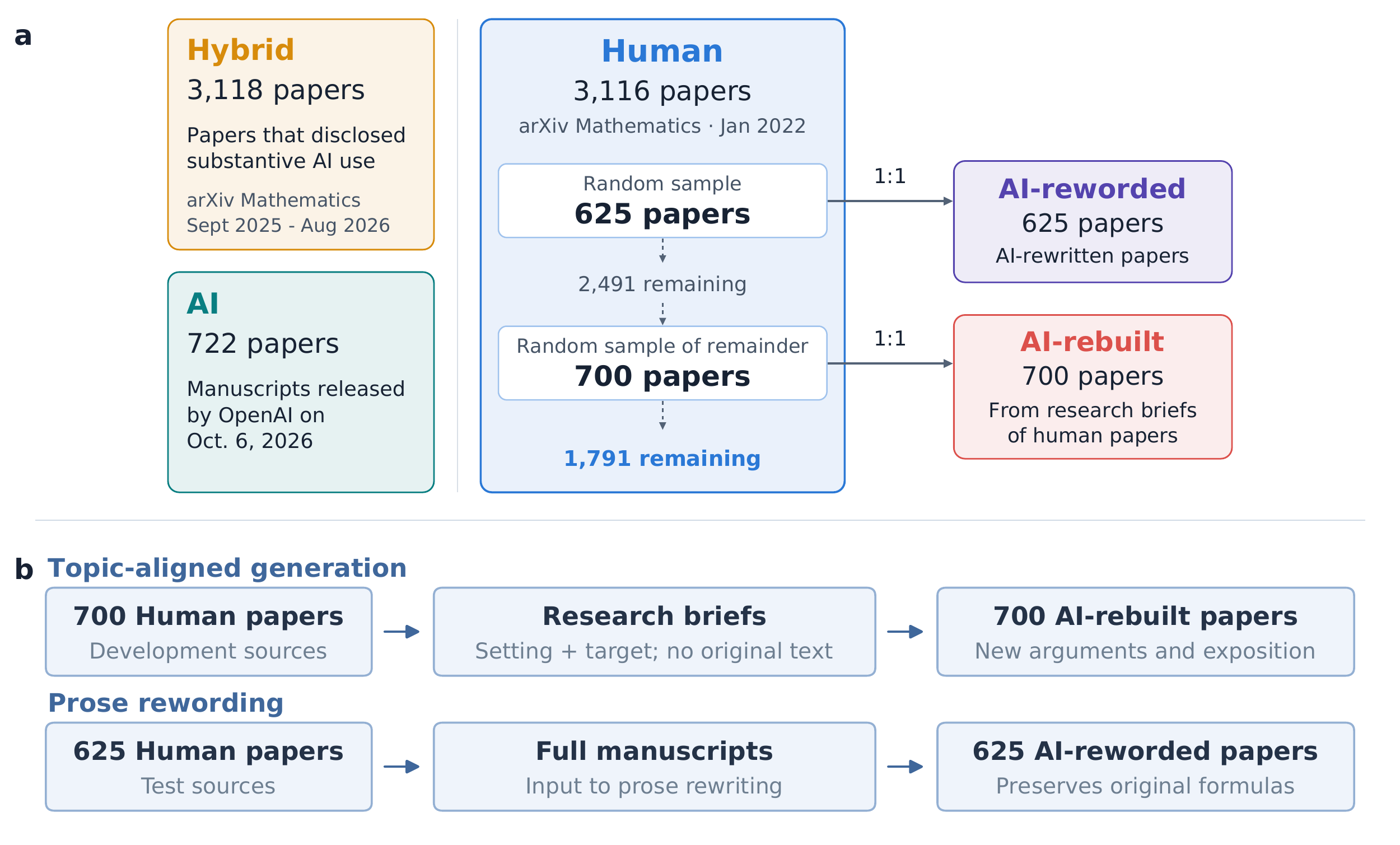}
\caption{(a) Five cohorts and study design. Sizes of the five cohorts and sampling of the Human cohort. (b) Two one-to-one pairing designs. Generation of the AI-rebuilt and AI-reworded papers from their Human sources.}
\label{fig:pairing-design}
\end{figure}

To study AI involvement in proof development, we need papers spanning different levels and forms of AI contribution. We also need comparisons that reduce confounding by research topic and help distinguish changes in mathematical organization from changes in linguistic presentation. To address these needs, we construct five cohorts by combining existing papers with manuscripts produced using generative AI tools (see Figure~\ref{fig:pairing-design} for an overview).

Three cohorts are drawn from existing sources: \emph{Human}, \emph{Hybrid}, and \emph{AI}. The Human cohort consists of arXiv mathematics papers from January 2022, providing a historical reference before widespread AI use in mathematical reasoning. The Hybrid cohort contains papers whose authors disclose \emph{substantive} AI assistance with mathematics, including contributions to ideas, proof development, checking, or formalization. Identifying these papers requires interpreting disclosure statements to distinguish contributions to mathematical content from non-substantive uses such as prose editing. We adopt the approach in \citet{JinKeSui2026} to prepare the disclosure labels through a combination of LLM assessment and manual auditing. The AI cohort comprises 722 mathematical manuscripts released by OpenAI and produced by its internal model with minimal human involvement \citep{OpenAI2026MathRelease}. Data collection and the preparation of the disclosure labels are described in \citet{JinKeSui2026}, and Appendix~\ref{app:generation} provides further details on data processing.

The AI cohort provides a reference set of purely AI-generated papers. However, its research topics may differ substantially from those of human-written papers. To mitigate confounding by research topics, we seek matched human-written and AI-generated paper pairs addressing the same topic. For a given human-written paper, our goal is to approximate a counterfactual: what might this paper look like if AI studied the same research topic? To pursue this goal, we use a specially designed paper-generation pipeline.

This pipeline produces the \emph{AI-rebuilt} cohort (Figure~\ref{fig:pairing-design}b). For each of 700 Human papers, we construct a research brief specifying its mathematical setting and target conclusion. GPT-6 Astra receives the brief without the original paper text and generates a full manuscript, including arguments and exposition. The resulting 700 Human--AI-rebuilt pairs align research settings and target conclusions, while the generated proofs, intermediate results, and other mathematical details can be different.

Here is one example of the Human--AI-rebuilt paper pair. The Human paper \citep{bridges2022weighted} is titled \emph{Weighted Cylindric Partitions}; its AI-rebuilt counterpart has a longer title: \emph{Weighted Cylindric Partitions and Double Shifted Plane Partitions: Functional Relations and Sum--Product Identities}. Both study generating functions and sum--product identities for related partition structures, including symmetric cases of widths 4 and 6. The Human paper emphasizes particular identities and applications; the AI-rebuilt manuscript develops general admissibility and convergence results, a positive-sum formula, and a finite $q$-difference system before treating these cases. It states two kernel lemmas, whereas the Human paper has one explicitly labeled lemma for a width-4 calculation. Thus, the pairing aligns the topic and mathematical aims without reproducing the same intermediate statements or proof sequence.

In the Hybrid, AI, and AI-rebuilt cohorts, AI involvement may affect mathematical reasoning and organization as well as written presentation. To help distinguish these effects, we add a reference cohort whose mathematical content originates with human authors but whose prose is extensively reworded by AI. The intervention aims to change the writing without asking AI to develop new mathematical content.

Specifically, to create the \emph{AI-reworded} cohort (Figure~\ref{fig:pairing-design}b), an LLM receives each of 625 complete Human test papers and is instructed to rewrite the prose while preserving equations, formal statements, citations, and reasoning steps. Each output is paired with its original to assess sensitivity to linguistic presentation and help distinguish it from structural differences associated with mathematical generation or assistance. 
Details are deferred to Appendix~\ref{app:rewording}. 

The 625 Human test papers are sampled from the full Human cohort; the 700 Human--AI-rebuilt pairs use a separate sample from the remaining 2,491 Human papers (Figure~\ref{fig:pairing-design}a). Thus, the two paired datasets use disjoint Human source sets. Together, the five cohorts contain 8,281 distinct papers. 

Although manuscripts in three of the five cohorts are produced using AI, the nature of AI involvement differs across these cohorts. In the AI-reworded cohort, the mathematical content originates from human-written papers. Thus, for our study of AI involvement in proof development, the Human and AI-reworded cohorts serve as controls, while the AI and AI-rebuilt cohorts serve as cases. The Human--AI-rebuilt pairs provide matched case-control comparisons.

\section{The ProofTrace features}
\label{sec:prooftrace-framework}

Feature construction is another central challenge in our study. Once papers are represented by suitable features, a wide range of statistical and machine-learning methods can be applied. A closely related body of work concerns AI-generated-text detection (see Section~\ref{sec:detection}), which uses token statistics, neural text representations, and scores or embeddings from pretrained language models as features. These features characterize writing but do not directly describe dependencies among mathematical results. We therefore develop ProofTrace, a feature construction pipeline that first represents each paper as a directed, attributed graph of mathematical results and their dependencies, then extracts interpretable features from this representation.

\subsection{From a mathematical paper to the GIPS representation}\label{subsec:GIPS}

Given a mathematical paper, its \emph{graph of immediate proof support} (GIPS) is a directed graph whose nodes represent mathematical results. A node can be an explicitly stated or cited result, such as a theorem or lemma, or a substantive intermediate conclusion reached within a proof, such as an important displayed equation or an important argument mentioned in the proof text. An edge $A\to B$ records that the written argument establishing $B$ directly uses the conclusion of $A$. We retain isolated results and disconnected components rather than assuming that every result belongs to a single proof chain.

The \emph{full GIPS representation} combines the graph with node and edge attributes. Node attributes record result type, statement and proof locations, external attribution, a content summary, and a supporting quotation. Edge attributes locate and quote a premise's use in the target argument (Appendix~\ref{app:gips-attributes}). 

To construct the full GIPS representation, we first preprocess each paper to obtain its Markdown text, and then use a large language model (LLM), guided by carefully designed prompts, to extract the graph and its attributes.

A conservative strategy would restrict nodes to explicitly stated results. This simplifies extraction by avoiding the additional judgment needed to identify substantive intermediate conclusions. However, the resulting graphs are often too sparse to capture the reasoning within long proofs, limiting their usefulness for downstream analysis. We therefore include substantive intermediate conclusions as nodes and devote substantial effort to making their identification more consistent.

We developed the extraction protocol through pilot runs, manual inspection, and iterative revisions to the prompts. A parser first anchors explicitly stated results in the source text, after which GPT-5.6 Luna processes the paper in successive portions rather than all at once. The extraction has two stages: a node stage that identifies the intermediate nodes, and an edge stage that identifies the direct-use edges. Each stage also records its supporting source evidence. We apply the same protocol across all five cohorts (Appendix~\ref{sec:response_schema}).

\begin{figure}[htbp]
\centering
\includegraphics[width=.84\linewidth]{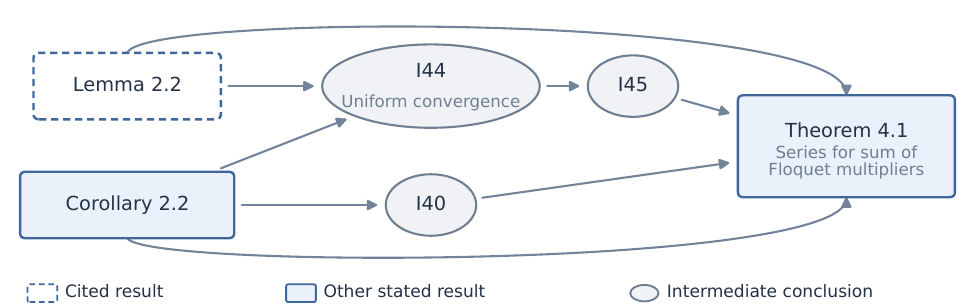}
\captionsetup{skip=4pt}
\caption{A six-node subgraph of the GIPS excerpt from \citet{WuXiaXu2022Floquet}. Arrows show direct use; I40, I44, and I45 are within-proof conclusions. Features use the complete 77-node graph (Figure~\ref{fig:app-real-rdg}).}
\label{fig:workflow}
\end{figure}

We illustrate the GIPS representation using the paper by \citet{WuXiaXu2022Floquet}. Its extracted graph contains 77 nodes and 77 edges. The nodes comprise 13 stated results, including three attributed to external sources, and 64 intermediate conclusions. Figure~\ref{fig:workflow} shows a six-node subgraph. The complete 77-node graph, along with examples of node/edge attributes,  are provided in Appendix~\ref{app:rdg-worked-example}.

There exist related works that use dependency graphs to represent the logical organization of mathematical papers. Graphs derived from formal Lean declarations, such as mathlib, capture dependencies among formalized results \citep{LiEtAl2026MathlibNetwork}; however, most papers in our cohorts have no Lean formalization, and formalizing them would require substantial effort. TheoremGraph \citep{KurganEtAl2026TheoremGraph} constructs graphs directly from non-formalized papers, but its graph contains only stated theorem-like results, excluding conclusions developed within proofs. It therefore omits much of the within-proof structure needed for our later findings about intermediate conclusions and their reuse. ProofTrace represents those conclusions as nodes and records their direct uses, making that structure available for feature construction.

\subsection{From the GIPS to interpretable structural features}
\label{subsec:feature-design}

We constructed features from GIPSs through a human--AI collaborative process. Drawing on mathematical intuition and preliminary inspection of the graphs, we first identified broad directions for feature design. An LLM then proposed candidate features within these directions. We manually reviewed each candidate for interpretability, mathematical well-definedness, and computability from the GIPS representation, removing unsuitable candidates. We also refined the feature definitions, for example by normalizing most count features by appropriate scales to reduce sensitivity to graph size. This process yielded 342 features (Appendix~\ref{sec:feature_registry}).

After manually reviewing their meanings, we grouped the features into six categories (Table~\ref{tab:feature-families}). \emph{References} captures external attribution and the use of cited results, while \emph{Organization} describes the arrangement of results and proofs in the text. \emph{Node Statistics} summarizes the composition of stated results and intermediate conclusions, and \emph{Edge Statistics} characterizes their direct-use relationships. \emph{Graph Motifs} measures local dependency patterns, while \emph{Global Topology} describes overall connectivity, reachability, and separate argument strands. This grouping primarily aids interpretation and visualization of the results.

We use the same example in Figure~\ref{fig:workflow} to illustrate the constructed features. Its three externally cited nodes give a cited-result fraction of $3/77\approx3.9\%$, and its 64 intermediate conclusions give $64/13\approx4.92$ intermediate conclusions per stated result. These quantities describe explicit attribution and the granularity of the extracted reasoning. Connectivity provides complementary information: the largest weakly connected component contains 53 nodes, or $68.8\%$ of the graph. Local configurations distinguish ways of combining results; for example, Corollary~2.2 contributes to Theorem~4.1 both directly and through I40, forming a feed-forward triangle visible in Figure~\ref{fig:workflow}. Source locations additionally describe how the reasoning is arranged in the text. Eight dependencies in the complete graph have a premise whose first recorded position follows that of its conclusion; this measures order of exposition. See Table~\ref{tab:feature-families}.

\begin{table}[htbp]
\centering
\caption{Six feature families, one example per family. The `Count' column contains the total number of features in each family. The `Example value' column is calculated on the extracted GIPS of \citet{WuXiaXu2022Floquet}.}
\label{tab:feature-families}
\small
\renewcommand{\arraystretch}{1.16}
\setlength{\tabcolsep}{4pt}
\begin{tabularx}{\linewidth}{@{}>{\raggedright\arraybackslash}p{1.05in}r>{\raggedright\arraybackslash}X>{\raggedright\arraybackslash}p{1.04in}@{}}
\toprule
Family & Count & Example measurement & Example value \\
\midrule
References & 38 & Cited-result nodes / all nodes & $3/77\approx0.039$ \\
Organization & 82 & Fraction of dependencies whose premise first appears later in the text & $8/77\approx0.104$ \\
Node Statistics & 28 & Intermediate conclusions / stated results & $64/13\approx4.92$ \\
Edge Statistics & 86 & Maximum in-degree / mean in-degree & $8/(77/77)=8$ \\
Graph Motifs & 34 & Feed-forward triangles / directed two-step walks & $1/95\approx0.011$ \\
Global Topology & 74 & Nodes in the largest weakly connected component / all nodes & $53/77\approx0.688$ \\
\bottomrule
\end{tabularx}
\end{table}

We call these features \emph{structural} because they use dependency relations, result types, external attribution, and source locations, without using the wording of recorded quotations or summaries as linguistic predictors. For comparison, we separately constructed 37 \emph{linguistic features} from the text recorded in the node and edge attributes of the GIPSs. These features are primarily intended to complement the structural features and improve prediction of AI involvement when AI assists with both proof development and writing (see Section~\ref{sec:detection}).

\section{Analyses and Discoveries}
\label{sec:analyses}

Using the constructed five cohorts and ProofTrace features, we have conducted various statistical analyses. Section~\ref{sec:landscape} focuses on exploratory data analysis and the overall trend, Section~\ref{sec:signatures} contains feature discovery and feature selection, and Section~\ref{sec:detection} builds detectors for detecting AI-assisted proof development.  

\subsection{The landscape of AI involvement}
\label{sec:landscape}

We compare the five cohorts as a whole in the ProofTrace feature space. The goal is to examine whether this space reflects different degrees of AI involvement in mathematics while remaining relatively stable when Human papers are reworded by AI. Here we use descriptive summaries, including principal-component projections and cohort means. 
Specifically, Section~\ref{subsec:landscape-gradient} shows a qualitative gradient across the five cohorts, and Section~\ref{subsec:temporal-landscape} documents a temporal shift in the Hybrid cohort's structural profile. 

\subsubsection{A qualitative gradient}
\label{subsec:landscape-gradient}

The five cohorts described in Section~\ref{sec:cohort-design} span different levels and forms of AI involvement. Human papers provide the reference; AI-reworded papers use AI to revise prose while aiming to preserve human mathematics; Hybrid papers report varying substantive mathematical assistance; and AI-rebuilt and AI papers are generated by AI. 
Therefore, a qualitative measure of the degree of {\it AI-assistance in proof development} indicates that
\begin{equation} \label{qualitative-gradient}
\mbox{Human}\approx \mbox{AI-reworded} < \mbox{Hybrid} < \mbox{AI-rebuilt and AI}. 
\end{equation}
We demonstrate that ProofTrace features reflectly such a qualitative gradient, as we have hoped.

\begin{figure}[t]
\centering
\includegraphics[width=\linewidth,height=.54\textheight,keepaspectratio]{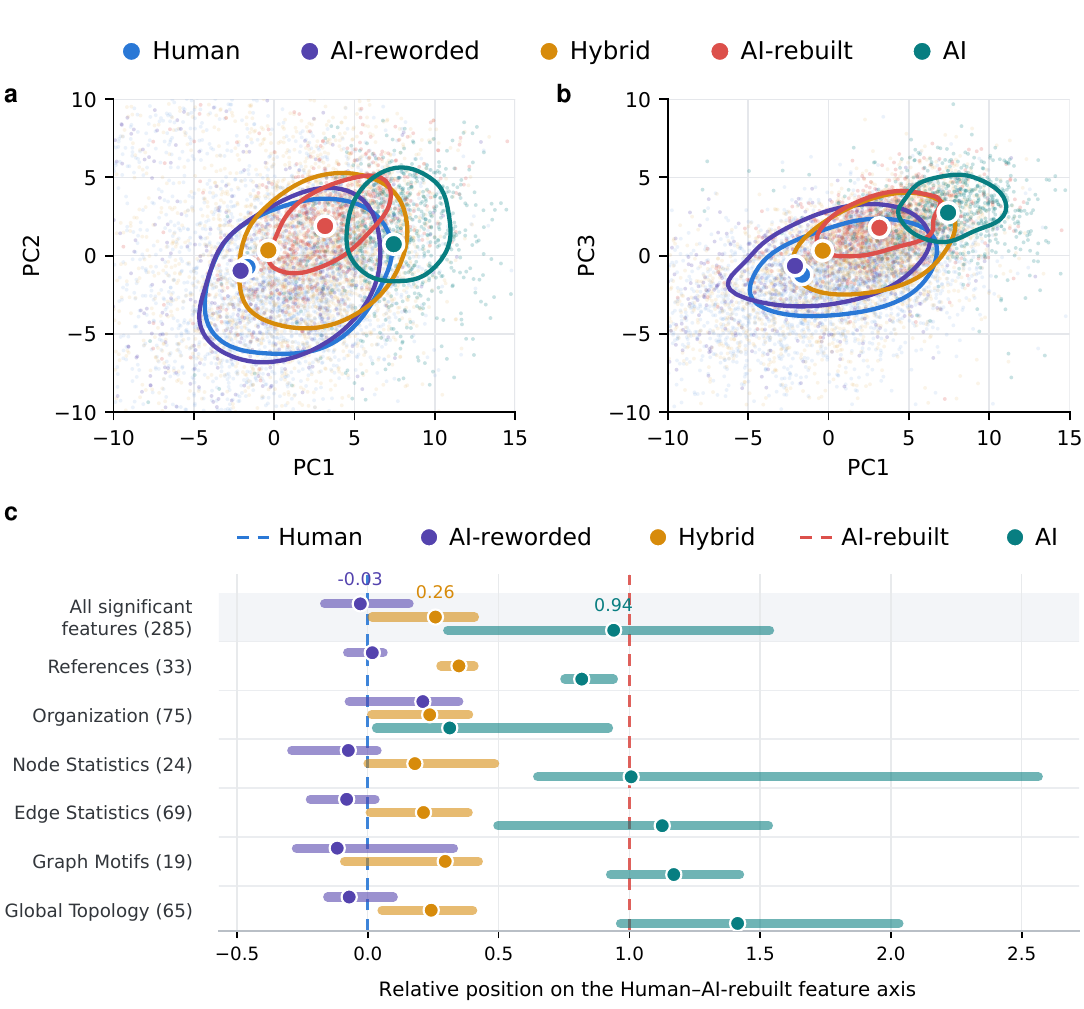}
\caption{Structural landscape of the five cohorts. (a,b) PC1--PC2 and PC1--PC3 projections. 8,281 papers on 342 structural features. Contours enclose approximately the densest half of each cohort, and larger dots show cohort means. (c) Relative positions of cohort means. Computed on the 285 features with a significant unpaired difference between Human ($r=0$) and AI-rebuilt ($r=1$). Dots and bars show medians and interquartile ranges.}
\label{fig:exp4pf-landscape}
\end{figure}

First, we examine principal-component projections of all 8,281 documents (Figure~\ref{fig:exp4pf-landscape}a,b). Human and AI-reworded papers overlap broadly; Hybrid papers occupy nearby but shifted regions; and AI-rebuilt and AI papers depart further. These observations are consistent with the expected qualitative gradient.

Notably, AI papers, produced by OpenAI's unreleased internal models, lie farther from Human papers than AI-rebuilt papers along PC1. The AI-rebuilt cohort also shifts away from Human papers along PC3. Research topics and generation conditions differ, so we cannot attribute the PC1 contrast to the models alone. Nevertheless, the manuscripts produced by the internal OpenAI models and those generated with GPT-6 Astra have visibly different positions in this feature space.

Next, to better study the direction of these shifts across individual features, we introduce a relative-position coordinate. For each feature $j$ and each cohort $k\in \{$ `Human', `AI-reworded', `Hybrid', `AI-rebuilt', and `AI'$\}$, let $m_{k,j}$ be the sample mean of feature $j$ within cohort $k$, and define
\begin{equation} \label{def:r}
r_{k,j}=\frac{m_{k,j}-m_{\mathrm{Human},j}}{m_{\mathrm{AI\text{-}rebuilt},j}-m_{\mathrm{Human},j}}.
\end{equation}
Human and AI-rebuilt cohorts occupy positions 0 and 1 for all features, respectively. For a fixed cohort of interest, these summary statistics allow us to see whether each feature moves toward AI-rebuilt, away from it, or beyond it. To limit noise from features without a reliably detected AI-rebuilt--Human difference, we focus on the 285 features with a significant difference in the full-data, unpaired comparison (two-sided tests with Benjamini--Hochberg-adjusted $q<0.05$; Appendix~\ref{app:statistical-specifications}).

The feature-wise positions broadly reinforce the pattern seen in the PCA plots (Figure~\ref{fig:exp4pf-landscape}c). AI-reworded papers remain close to Human papers (median $r=-0.03$); Hybrid papers occupy an intermediate position (median $r=0.26$); and AI papers depart much further from Human papers (median $r=0.94$). 
The AI cohort moves in the AI-rebuilt direction on 243 of the 285 features and extends beyond AI-rebuilt on 132, although its overall median remains near the AI-rebuilt anchor. 

The feature-wise positions also makes a second finding quantitative: the Hybrid cohort lies only about one-quarter of the way from Human to AI-rebuilt. At the cohort level, this suggests that papers disclosing substantive AI use still look more like human-written papers than AI-generated papers, when measured by their ProofTrace feature values.



\subsubsection{Evolvement of the Hybrid cohort}
\label{subsec:temporal-landscape}

The feature-wise position statistics in \eqref{def:r} provide a way to measure where Hybrid papers lie between the Human and AI-rebuilt references. Recall that all Hybrid papers are arXiv submissions posted from September 2025 to August 2026.
Section~\ref{subsec:landscape-gradient} computed these statistics using all these papers. We now investigate the trend over time.

\begin{figure}[tbp]
\centering
\includegraphics[width=.95\linewidth,height=.4\textheight,keepaspectratio]{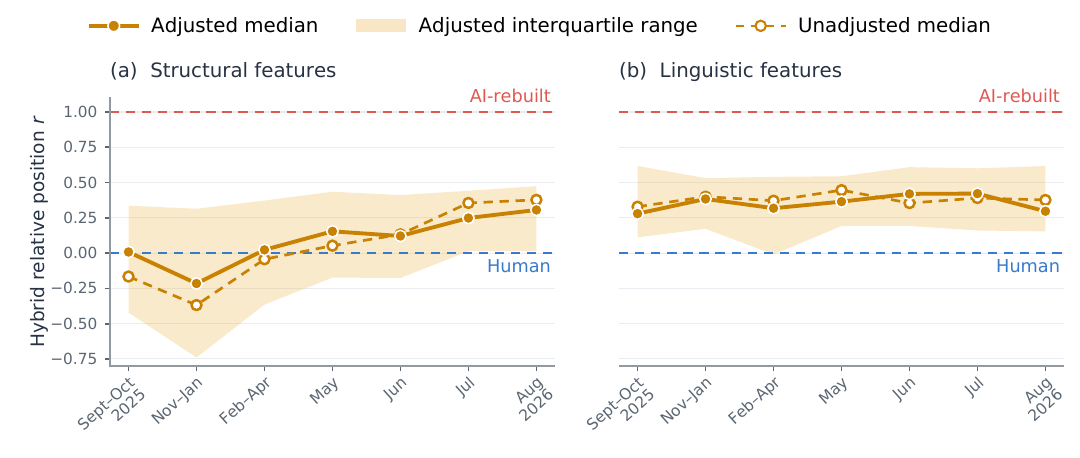}
\caption{Temporal positions of Hybrid papers from September 2025 to August 2026 relative to Human ($r=0$) and AI-rebuilt ($r=1$) reference means. Panels show (a) 265 structural and (b) 35 linguistic features with significant paired AI-rebuilt--Human differences. Solid lines: medians after adjusting period means for research field, reported AI-use group and revised-version status. Dashed lines: unadjusted medians. Shaded bands: interquartile ranges across features.}
\label{fig:disclosed-temporal-positions}
\end{figure}

We divide the twelve months into seven periods. For each feature $j$ and period $p$, we use a least-squares model to estimate the Hybrid mean $m_{jp}$, standardizing the mix of research fields, reported AI-use groups, and manuscript versions to the pooled Hybrid papers with that feature. With $h_j$ and $a_j$ denoting the Human and AI-rebuilt means over complete source pairs, we compute $r_{jp}=(m_{jp}-h_j)/(a_j-h_j)$. See Appendix~\ref{app:temporal-variation}. We select the 265 structural and 35 linguistic features with statistically significant paired AI-rebuilt--Human differences, then take the median and interquartile range of $r_{jp}$ across structural features and, separately, across linguistic features in each period (Figure~\ref{fig:disclosed-temporal-positions}). Figure~\ref{fig:disclosed-temporal-positions} also shows medians computed without adjustment. These are period-level summaries across features, not positions assigned to individual papers.

The adjusted structural median moves from $0.01$ in September--October 2025 to $0.31$ in August 2026, after dipping to $-0.22$ in November 2025--January 2026 (Figure~\ref{fig:disclosed-temporal-positions}a). Of the 265 structural features, 187 have higher adjusted positions at the end than at the start. The unadjusted median rises from $-0.17$ to $0.38$; adjustment reduces but does not remove the shift. In contrast, the adjusted linguistic median changes from $0.28$ to $0.30$, without a comparable aggregate trend (Figure~\ref{fig:disclosed-temporal-positions}b), although individual features may change. 

These descriptive patterns cannot distinguish changes in model capabilities, AI-use practices, or selection into disclosure from other developments. Historical Human papers were posted in a different period, so we cannot assess whether contemporaneous Human papers changed similarly. 

\subsection{Significant features}
\label{sec:signatures}

Section~\ref{sec:landscape} shows that ProofTrace features distinguish the cohorts in our study. We now ask which individual features show statistically significant differences using the paired Human--AI-rebuilt design. Each AI-rebuilt paper is generated from a brief specifying the Human paper's research setting and target conclusion, providing an approximate counterfactual for how the same research task might be developed by AI. We then validate these features using data from all cohorts. 

\subsubsection{The Human vs. AI-rebuilt comparison}

For each structural feature, we compare its values in the 700 Human--AI-rebuilt source pairs. A two-sided paired $t$-test is applied to the within-pair differences $D_i=x_{\mathrm{AI\text{-}rebuilt},i}-x_{\mathrm{Human},i}$, using pairs for which both values are defined. We report the signed standardized mean difference $d$, positive when the AI-rebuilt mean is higher. To account for testing many features, we apply the Benjamini--Hochberg procedure jointly across the structural features and call a mean difference significant when its adjusted $p$-value $q<0.05$. 
A supplementary two-sample Kolmogorov--Smirnov test compares the marginal distributions, with its own Benjamini--Hochberg adjustment (Appendix~\ref{app:statistical-specifications}).

\begin{figure}[tbp!]
\centering
\includegraphics[width=\linewidth]{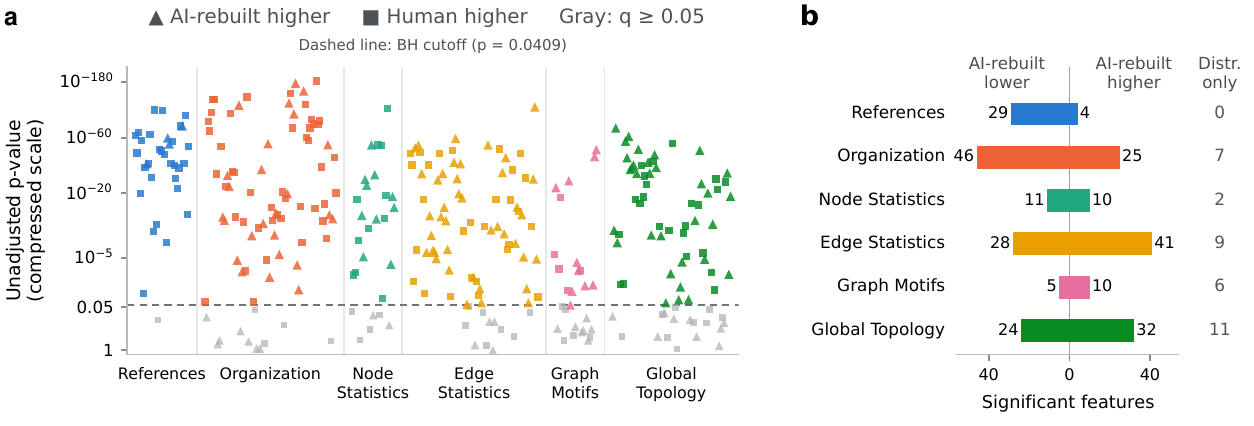}
\caption{(a) Manhattan-style display of structural differences. Paired-test results for 700 paired Human and AI-rebuilt papers, by family. Evaluation metric: unadjusted $p$-value, significant at Benjamini--Hochberg $q<0.05$. (b) Significant features per family. Results for the 265 features with a significant mean difference and the 35 additional features that differ only in distribution. }
\label{fig:exp1}
\end{figure}

Figure~\ref{fig:exp1} shows significant mean differences in 265 of the 342 structural features, spanning all six categories. Organization (71 of 82) and Edge Statistics (69 of 86) contribute the largest counts, while References has the highest proportion (33 of 38). The median $|d|$ across features is 0.38; another 35 features differ in distribution without a significant mean difference. These counts show the breadth of measured variation, as the differences between AI-rebuilt and Human papers are not confined to a few features or to one category, but extend across most structural features and all six categories.

\begin{figure}[t]
\centering
\includegraphics[width=\linewidth]{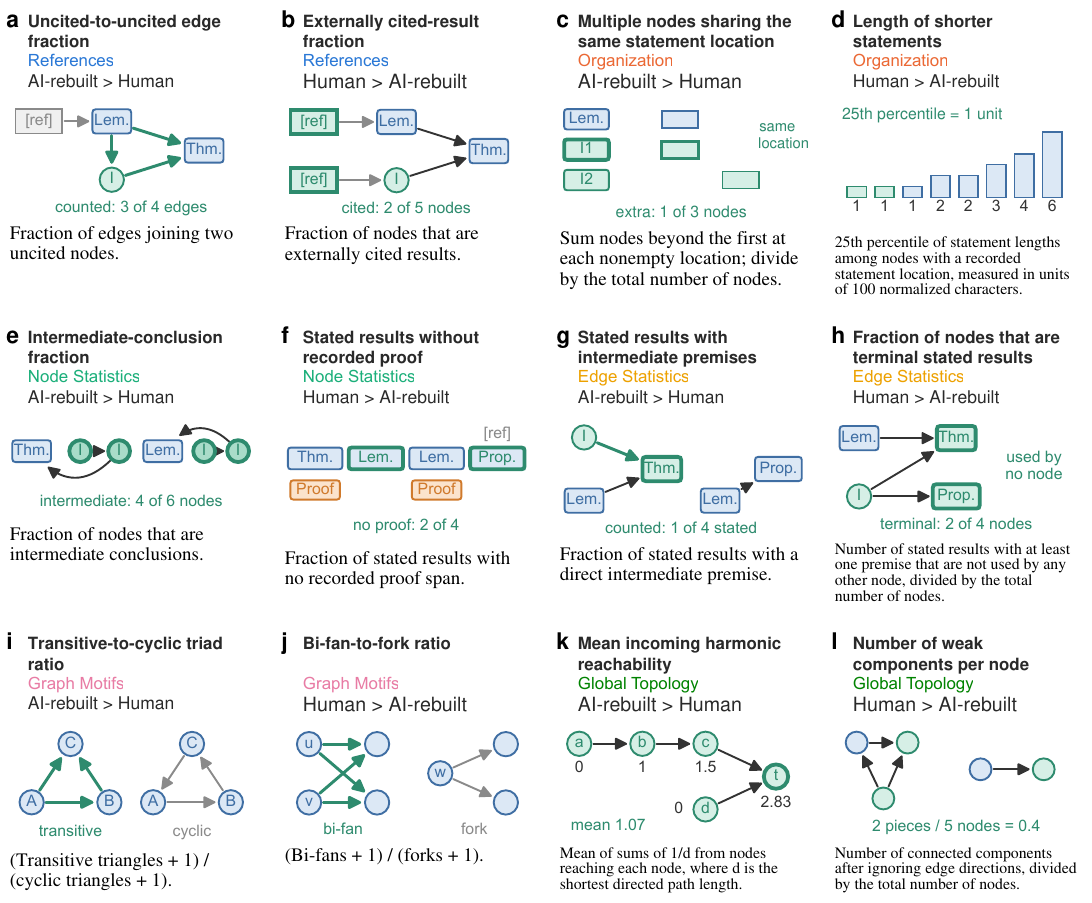}
\caption{Interpretations of twelve structural features in the paired Human--AI-rebuilt comparison. Each feature has largest or smallest standardized mean differences among significant features in the corresponding feature family. }
\label{fig:signature-interpretations}
\end{figure}

To interpret these differences, Figure~\ref{fig:signature-interpretations} selects, within each category, the significant feature with the largest positive AI-rebuilt--Human value of $d$ and the one with the most negative value. These examples illustrate distinct measurements. Together, they suggest four patterns in the extracted arguments.

\begin{figure}[t]
\centering
\includegraphics[width=\linewidth]{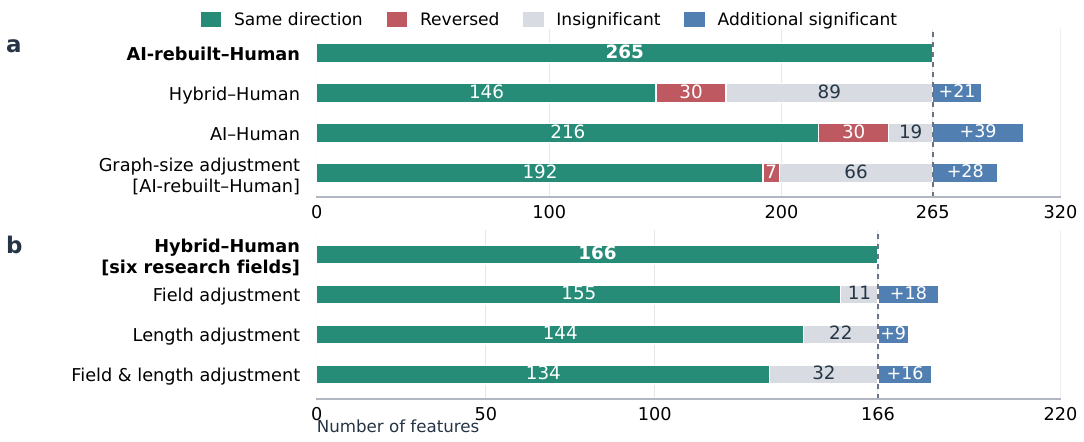}
\caption{Recurrence and robustness of structural features. Reference rows show 265 significant AI-rebuilt–Human features in (a) and 166 significant Hybrid–Human features restricted to six research fields in (b). Subsequent bars partition each reference set into significant features in the same direction (green), significant reversals (red), and insignificant features (gray). Blue features are
outside the reference set and have a significant difference. Significance: Benjamini--Hochberg $q<0.05$.}
\label{fig:persistence}
\end{figure}

\noindent\emph{Grouping conclusions.} Distinct nodes in AI-rebuilt graphs more often share the same recorded statement location (0.084 versus 0.018 per node; Figure~\ref{fig:signature-interpretations}c), while the 25th-percentile statement length is shorter (3.21 versus 3.99 units in each text chunk; panel d). The shared locations are consistent with several extracted conclusions occupying one passage; shorter statement lengths are a separate aspect of organization. 

\noindent\emph{Attribution to prior work.} Almost all AI-rebuilt dependencies join uncited results (99.4\% versus 89.9\% in Human papers), whereas externally cited results account for just 0.2\% of AI-rebuilt nodes versus 6.1\% of Human nodes (panels a,b). Human papers also more often contain stated results without a recorded proof span (21\% versus 0.2\%; panel f). The first two features indicate more explicit use and credit of outside mathematical results in the extracted Human arguments.

\noindent\emph{Intermediate steps and reuse.} Intermediate conclusions make up 90\% of AI-rebuilt nodes versus 84\% of Human nodes (panel e) and serve as direct premises of 87\% versus 58\% of stated results (panel g). AI-rebuilt also has a larger transitive-to-cyclic triad ratio, driven mainly by transitive triangles (panel i), and higher incoming harmonic reachability (3.17 versus 2.21; panel k), indicating more connected upstream dependencies. Human graphs have a larger local bi-fan-to-fork ratio (panel j), consistent with shared pairs of premises feeding multiple results. Together, these differences suggest more in-proof development in AI-rebuilt papers and more reuse of shared premises in Human graphs. 

\noindent\emph{Separate strands.} Human graphs have more terminal stated results with at least one premise per node (0.038 versus 0.016; panel h) and more weakly connected components per node (0.24 versus 0.14; panel l). This pattern is consistent with more distinct endpoints and strands in the extracted Human arguments, which conveys auxiliary insights or connections to prior work.

Together, these twelve examples characterize the organization of recorded arguments. The effect of graph-size adjustment is examined in Section~\ref{subsec:controlled-comparisons}.

\subsubsection{Recurrence and robustness}
\label{subsec:controlled-comparisons}

Many signatures recur beyond the AI-rebuilt generation pipeline. Of its 265 significant differences from Human sources, 146 are also significant in the same direction in Hybrid papers and 216 in AI papers (Figure~\ref{fig:persistence}a). Recurrence in a separate generated collection and in papers reporting substantive assistance supports a common component of structural variation across these settings.

The departures are not uniform. Thirty features show significant reversals in each comparison; another 89 are insignificant in the Hybrid comparison and 19 in the AI comparison. Beyond the initially significant AI-rebuilt--Human features, another 21 features are significant in the Hybrid--Human comparison and 39 in the AI--Human comparison (Figure~\ref{fig:persistence}a). A lack of significance does not establish equality, whereas reversals identify dimensions along which the cohorts differ in their relationship to Human papers. The shared directions and reversals together suggest that AI involvement is associated with recurring organizational tendencies, rather than a single structural template. The present design does not determine whether a particular reversal arises from the model, the generation instructions, the mathematical task, or the form of human assistance.

We next assess whether these structural differences persist when accounting for research field, manuscript length, and graph size in the relevant comparisons. Field composition does not account for most of the significant Hybrid--Human differences in the six-field analysis. Before adjustment, 166 of the 342 structural features show significant differences; these 166 features form the common baseline for the field and length controls, independently of the AI-rebuilt--Human baseline in Figure~\ref{fig:persistence}a. Of these, 155 remain significant in the same direction after field adjustment, 144 after length-only matching, and 134 after matching on both research field and manuscript length. None of the three analyses has a significant reversal; 11, 22, and 32 differences, respectively, are insignificant. An additional 18, 9, and 16 features become significant under the corresponding controls (Figure~\ref{fig:persistence}b). Both matched draws contain the same number of papers per cohort (Appendix~\ref{app:length-matching}), and the length-only draw yields 153 significant structural differences in total. Agreement with the AI-rebuilt profile also remains after field adjustment. Using disjoint Human reference sets, 249 of the 342 structural features agree in direction between the two contrasts before adjustment and 253 afterward (Appendix~\ref{app:field-agreement}).

Within-field results qualify this aggregate consistency. Structural differences are significant in combinatorics, number theory, probability, and algebraic geometry, but none reaches significance in machine learning or optimization and control under the corresponding tests. The findings support recurring signatures while leaving room for variation in their strength across research areas (Appendix~\ref{app:field-adjustment}).

Graph-size adjustment addresses a different comparison and a different explanation. In the paired AI-rebuilt--Human analysis, adjustment for node and edge counts (Appendix~\ref{app:graph-size}) leaves 192 of the original 265 differences significant in the same direction, with a median adjusted-to-unadjusted effect ratio of 0.71; 7 show significant reversals and 66 are insignificant. A further 28 features become significant after adjustment (Figure~\ref{fig:persistence}a). 

\subsection{A detection experiment}
\label{sec:detection}

In this subsection, we present an application of our selected features to the detection of substantially AI-involved papers. This is a field in which people are increasingly interested, and we construct this experiment using the Human and Hybrid cohorts that we have collected. 

We construct the training set of our experiments using a random selection of 2,491 papers from the Human cohort and 2,489 papers from the Hybrid cohort. Then, on this training set, we train two classifiers of our own, which we call ProofTrace and ProofTrace++. ProofTrace uses only the 342 structural features that it computes from the GIPS of a paper. ProofTrace++ uses a selection of the structural features above and linguistic features, which it computes from the GIPS and the full text of a paper. 

To evaluate our methods, we conduct two experiments. For our first experiment, we investigate how our methods and the baselines perform. For this experiment, the testing set is the remaining 625 Human papers and the 629 Hybrid papers. We evaluate every method on this testing set, and for every method that requires training, we train it on the training set. We use the area under the receiver operating characteristic curve (AUROC) and average precision (AP) as evaluation metrics, with Hybrid papers as the positive class.

For our second experiment, we want to test whether or not these methods can really tell papers apart by their proof organization, rather than relying solely on linguistic features. The motivation is that, when a paper has substantial AI involvement, it is likely that AI also contributes to the wording of that paper. Therefore, we create a controlled experiment by constructing the testing set using the 629 Hybrid papers, and the 625 AI-reworded versions of the Human papers. Every method is then evaluated on this new testing set, using AUROC and AP as evaluation metrics.

In both experiments, we compare ProofTrace and ProofTrace++ with 10 baselines. The first five in Table~\ref{tab:detection} are pretrained and zero-shot baselines, and are not fine-tuned on our data. They are MELD \citep{li2026meld}, ModernBERT-MAGE, which is the ModernBERT detector of \citet{drayson2025collapse} trained on the MAGE corpus \citep{li2024mage}, DETree \citep{he2025detree}, Threads of Subtlety \citep[ToS,][]{kim2024threads} and RAIDAR \citep{mao2024raidar}. The other five are fitted and adapted baselines, and are fine-tuned on our training set. Each starts from a pretrained detector, keeps its encoder frozen and adapts the rest to our data. They are two variants of ModernBERT-MAGE, labeled (LR) and (finetuned), RADAR \citep{reword_radar}, Desklib \citep{reword_desklib_card}, which uses the DeBERTa-v3 encoder \citep{he2023debertav3}, and a variant of DETree labeled (kNN). Details of all these methods can be found in Appendix~\ref{app:release-benchmark}.

\begin{table}[t]
\centering
\caption{Detection results of the two experiments for ProofTrace, ProofTrace++ and the other baselines. Evaluation metric: AUROC and AP with Hybrid as positive, higher is better.}
\label{tab:detection}
\small
\setlength{\tabcolsep}{6pt}
\renewcommand{\arraystretch}{1.12}
\begin{tabular}{@{}lrrrr@{}}
\toprule
& \multicolumn{2}{c}{Original} & \multicolumn{2}{c}{Reworded} \\
\cmidrule(lr){2-3}\cmidrule(l){4-5}
Method / variant & AUROC & AP & AUROC & AP \\
\midrule
\multicolumn{5}{@{}l}{\textbf{Pretrained and zero-shot baselines}} \\
MELD (pretrained) & 0.7904 & 0.7602 & 0.1618 & 0.3369 \\
ModernBERT-MAGE (pretrained) & 0.4513 & 0.4889 & 0.4149 & 0.4360 \\
DETree (pretrained) & 0.4220 & 0.4505 & 0.4183 & 0.4494 \\
ToS (pretrained) & 0.4128 & 0.4349 & 0.1930 & 0.3425 \\
RAIDAR (zero-shot) & 0.6477 & 0.6937 & 0.1305 & 0.3280 \\
\midrule
\multicolumn{5}{@{}l}{\textbf{Fitted and adapted baselines}} \\
ModernBERT-MAGE (LR) & 0.9554 & 0.9622 & 0.5529 & 0.5397 \\
ModernBERT-MAGE (finetuned) & 0.9327 & 0.9465 & 0.4817 & 0.4707 \\
RADAR (finetuned) & 0.9576 & 0.9620 & 0.5952 & 0.6916 \\
Desklib (finetuned) & 0.9191 & 0.9237 & 0.4809 & 0.5301 \\
DETree (kNN) & 0.8777 & 0.8890 & 0.3665 & 0.4636 \\
\midrule
\multicolumn{5}{@{}l}{\textbf{ProofTrace classifiers}} \\
ProofTrace & 0.8306 & 0.8383 & 0.8365 & 0.8424 \\
ProofTrace++ & 0.8903 & 0.9048 & 0.6246 & 0.6488 \\
\bottomrule
\end{tabular}
\par\vspace{6pt}
\noindent\begin{minipage}{\textwidth}
\footnotesize
\end{minipage}
\end{table}

The results of the two experiments can be found in Table~\ref{tab:detection}. From these results, we have two very interesting findings. First, the figures show that ProofTrace is stable under rewording, with AUROC and AP barely changing in the two experiments, while most other methods deteriorate substantially. Since the rewording experiment aims to change the wording while preserving the mathematical organization, this suggests that ProofTrace captures a signal of AI involvement in substantial mathematical reasoning, rather than relying solely on wording and writing style to distinguish papers.

Our second finding is that, both structural and linguistic signals are discriminative, and in the first experiment, ProofTrace++, which combines the two, reaches a higher AUROC and AP than ProofTrace. Remarkably, with only a few hundred features, ProofTrace++ achieves performance on original papers similar to that of the fine-tuned baselines, which are built on large pretrained transformer models.

\section{Discussion}

To study the signatures of AI involvement in mathematical proofs, we construct five cohorts with a total of 8,281 papers. We introduce ProofTrace, a feature extraction framework that represents each paper as a GIPS and extracts interpretable features from this graph. Using these cohorts and features, we obtain three main findings. First, the features show a qualitative gradient of AI involvement. AI-reworded papers remain close to Human papers, AI-rebuilt and AI papers differ from Human papers, and Hybrid papers occupy the middle, shifting towards the AI side through time. Second, many features differ significantly between paired Human and AI-rebuilt papers, and these differences recur in the Hybrid and AI cohorts and persist after adjustment for graph size, research field and manuscript length. Third, in the detection experiments, ProofTrace is stable under rewording, showing its capability of detecting mathematical reasoning pattern difference, whereas ProofTrace++, which combines structural and linguistic features, performs similarly to the fine-tuned baselines on original papers.

Our work has many potential impacts in different fields. First, our work provides a framework for estimating the proportion of real AI use in a field. Disclosure of AI use is often incomplete, and our methodology offers a way to address this problem, which could be extended to any field that relies on mathematical reasoning, such as theoretical physics, statistics and econometrics. Second, our work provides insight into how to make AI-generated proofs more tractable. The features that we proposed could also give a principled way of developing AI agents for this purpose, such as extracting reusable lemmas or regrouping intermediate conclusions while preserving the mathematical reasoning. Third, our work has potential in building an alarm system for AI involvement in proofs. This would be especially beneficial for journals with a high volume of submissions, and our method might be able to give reference or useful information that facilitates the reviewing process.

Finally, we discuss the limitations of our study and the future work that they suggest. The first limitation is that we do not take adversarial attacks into account. Future AI-generated proofs may exploit the features, and an interesting line of work would be to improve the features to take such instances into account. The second limitation is that we only study the mathematical organization and presentation of a paper, rather than its proof techniques. An interesting direction would be to further refine the features and relate structural features to independent assessments of mathematical quality.

\bibliographystyle{plainnat}
\bibliography{goldrush2}

\clearpage
\appendix
\startcontents[appendix]
\section*{Appendix contents}
\begingroup
\small
\printcontents[appendix]{}{1}[2]{}
\endgroup
\counterwithin*{figure}{section}
\counterwithin*{table}{section}
\renewcommand{\thefigure}{\thesection\arabic{figure}}
\renewcommand{\thetable}{\thesection\arabic{table}}
\renewcommand{\theHfigure}{\thesection.\arabic{figure}}
\renewcommand{\theHtable}{\thesection.\arabic{table}}

\section{Data construction}
\label{app:data}
\label{app:generation}

\definecolor{genblue}{rgb}{0.161,0.353,0.541}
\definecolor{gengreen}{rgb}{0.216,0.435,0.420}
\definecolor{genred}{rgb}{0.557,0.298,0.259}
\definecolor{genbluebg}{rgb}{0.949,0.965,0.980}
\definecolor{gengreenbg}{rgb}{0.945,0.969,0.965}
\definecolor{genredbg}{rgb}{0.984,0.957,0.945}
\newcommand{\genstep}[6]{\fcolorbox{#1}{#2}{\parbox[c][#5][c]{#6}{\centering\textcolor{#1}{\textbf{#3}}\par\vspace{4pt}#4}}}

\begin{figure}[H]
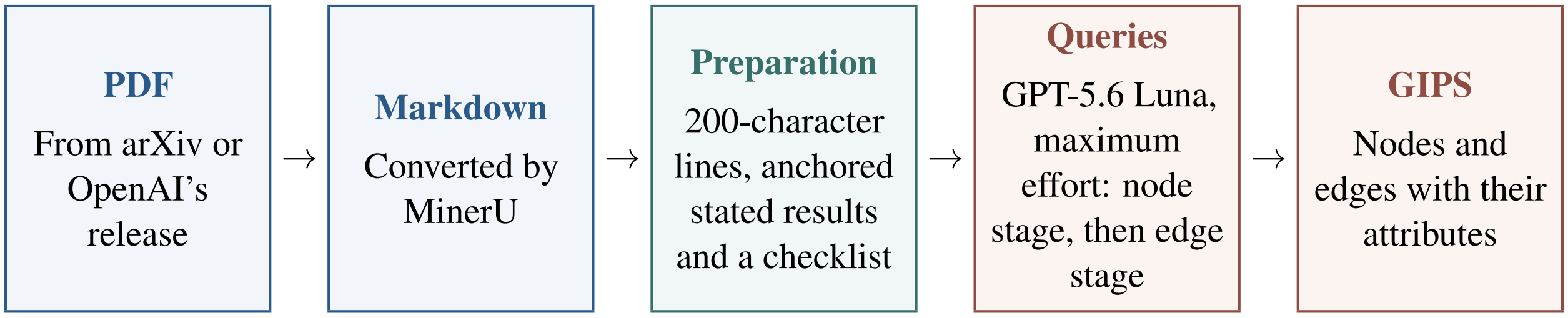

\centering
\setlength{\fboxrule}{0.8pt}%
\setlength{\fboxsep}{4pt}%
\genstep{genblue}{genbluebg}{PDF}{From arXiv or OpenAI's release}{1.15in}{0.98in}\hfill$\rightarrow$\hfill
\genstep{genblue}{genbluebg}{Markdown}{Converted by MinerU}{1.15in}{0.98in}\hfill$\rightarrow$\hfill
\genstep{gengreen}{gengreenbg}{Preparation}{200-character lines, anchored stated results and a checklist}{1.15in}{0.98in}\hfill$\rightarrow$\hfill
\genstep{genred}{genredbg}{Queries}{GPT-5.6 Luna, maximum effort: node stage, then edge stage}{1.15in}{0.98in}\hfill$\rightarrow$\hfill
\genstep{genred}{genredbg}{GIPS}{Nodes and edges with their attributes}{1.15in}{0.98in}
\caption{From the PDF of a paper to its GIPS.}
\label{fig:app-pipeline}
\end{figure}

Figure~\ref{fig:app-pipeline} shows how we turn the PDF of a paper into its GIPS. For the AI-rebuilt and AI-reworded papers, there is no PDF. An LLM writes their Markdown, and the remaining steps are the same.

\subsection{Human, Hybrid and AI papers}
\label{app:pipeline-converted}
\label{sec:response_schema}

First, we convert the PDF of each Human, Hybrid and AI paper to Markdown with MinerU \citep{Wang2024MinerU}. The Markdown text is then divided into lines of 200 characters for reference, and is used as the input of the extraction step. To help the LLMs locate and extract the theorems more clearly and with more stability, we use a parser to anchor the locations of the explicitly formatted theorems, lemmas, propositions, corollaries, claims, facts and observations in the Markdown file. Next, we use GPT-5.6 Luna with maximum reasoning effort to extract each GIPS in two stages. In each of these stages, we use a fixed prompt and require the model to output its graphs in a fixed JSON schema. The model reads the paper and writes its output through tool calls in the Codex CLI.

In the first (node) stage, the model finds the nodes. It keeps the anchored stated results as they are, locates their proofs, and adds an intermediate node for every other conclusion that the paper asserts as established. In a computation, only the final relation counts, unless a step is numbered, tagged or labeled, or the paper refers to it later. As a rule, one asserting sentence gives one node, but parts with their own number or label give their own nodes. Repeated statements of the same conclusion are merged into one node. A second parser also makes a checklist of passages that may assert a conclusion, and the model checks every listed passage with these rules.

In the second (edge) stage, the nodes are fixed and the model adds the edges. It draws an edge from node $A$ to node $B$ only if the argument for $B$ directly uses the conclusion of $A$. For each edge, it records all the places where this use happens and quotes the text that shows it. The model goes through the nodes in the order of the paper. In the argument for each node, it finds the node that each label, equation number or pointer such as ``the lemma above'' refers to. It also adds uses that carry no such mark but still leave a trace in the text.

\subsection{AI-rebuilt papers}
\label{app:pipeline-rebuilt}

\begin{figure}[H]
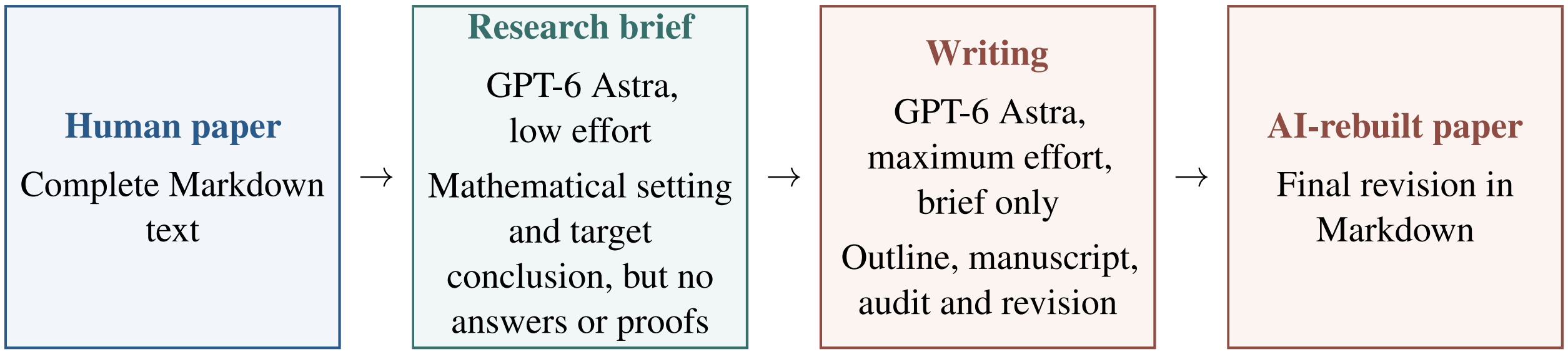

\centering
\setlength{\fboxrule}{0.8pt}%
\setlength{\fboxsep}{4pt}%
\genstep{genblue}{genbluebg}{Human paper}{Complete Markdown text}{1.3in}{1.27in}\hfill$\rightarrow$\hfill
\genstep{gengreen}{gengreenbg}{Research brief}{GPT-6 Astra,\\low effort\par\vspace{3pt}Mathematical setting and target conclusion, but no answers or proofs}{1.3in}{1.27in}\hfill$\rightarrow$\hfill
\genstep{genred}{genredbg}{Writing}{GPT-6 Astra,\\maximum effort,\\brief only\par\vspace{3pt}Outline, manuscript, audit and revision}{1.3in}{1.27in}\hfill$\rightarrow$\hfill
\genstep{genred}{genredbg}{AI-rebuilt paper}{Final revision in Markdown}{1.3in}{1.27in}
\caption{Construction of an AI-rebuilt paper from its Human source paper.}
\label{fig:app-generation}
\end{figure}

The pipeline for the AI-rebuilt papers is the same as in Appendix~\ref{app:pipeline-converted}, except that we do not convert a PDF. Instead, GPT-6 Astra writes the Markdown of each of the 700 AI-rebuilt papers from a research brief of its Human source paper (Figure~\ref{fig:app-generation}). First, we give GPT-6 Astra (low reasoning effort) the complete Markdown of the Human paper and ask it for a brief of one or two paragraphs, written for a mathematician who will not see the paper. In the paper's own terminology and notation, the brief describes the mathematical setting (the main object and the essential hypotheses) and states the target conclusion as a task to discover, define and prove. It does not give the answers, formulas, intermediate results, proof strategies or proof steps that solve this task. The brief is then checked against the Human paper.

Next, in a new conversation, GPT-6 Astra (maximum reasoning effort) writes a new paper from the brief alone. It is told not to browse the web or use other tools. Writing the paper takes three requests: first an outline, then a complete manuscript with definitions, statements and proofs, and finally an audit and revision of this manuscript. We take the final revision, in Markdown, as the AI-rebuilt paper.

\subsection{AI-reworded papers}
\label{app:rewording}

Similarly, the pipeline for the AI-reworded papers differs from the one in Appendix~\ref{app:pipeline-converted} only in how the Markdown is obtained. For each of the 625 Human test papers, GPT-5.6 Luna (extra-high reasoning effort) rewrites the Markdown in one request. Together with the Markdown, the model gets a ledger. The ledger lists what must be preserved (the equation spans, notation, formal statements, citations and propositions that carry the reasoning), and it marks the regions where the prose can change without breaking a dependency. We ask the model to rewrite nearly all of the remaining prose. It may merge, split or reorganize paragraphs within these regions, as long as every definition still comes before its first use. We also ask it not to add claims, examples, experiments, citations, assumptions or conclusions, and not to remove any reasoning step. The model returns only the rewritten Markdown, which is the AI-reworded paper.

\section{GIPS attributes}
\label{app:gips}

\subsection{Node and edge attributes}
\label{app:gips-attributes}

Tables~\ref{tab:rdg-node-schema} and~\ref{tab:rdg-edge-schema} list the attributes. Locations are inclusive ranges of the lines of Appendix~\ref{sec:response_schema}.

\begin{figure}[p]
\centering
\includegraphics[width=.95\linewidth,height=.9\textheight,keepaspectratio]{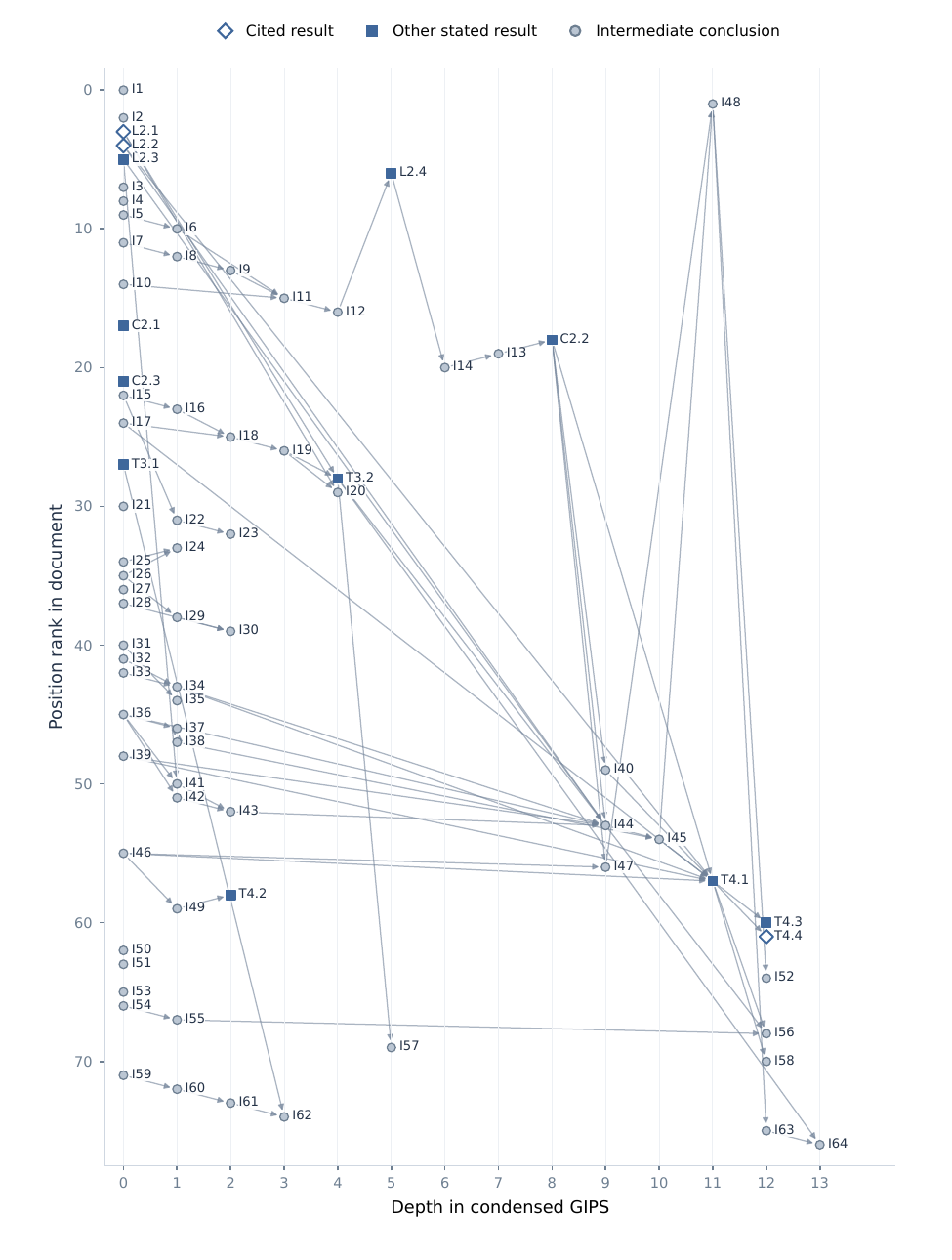}
\caption{The complete GIPS of \citet{WuXiaXu2022Floquet}. We place each node vertically by the first recorded position of its statement and horizontally by the number of edges on the longest chain of dependencies that ends at it, with I37 and I38, which use each other, counted as one node. L, C, and T stand for lemmas, corollaries, and theorems, and I stands for intermediate conclusions.}
\label{fig:app-real-rdg}
\end{figure}

\begin{table}[htb]
\centering
\caption{The attributes of each node.}
\label{tab:rdg-node-schema}
\begin{tabularx}{\linewidth}{@{}p{.27\linewidth}X@{}}\toprule
Attribute & Content\\\midrule
\texttt{id} & The parser's label of a stated result (e.g., Lemma~2.4, or U1, U2, \dots{} if it has no number). Intermediate conclusions are labeled I1, I2, and so on.\\
\texttt{kind} & \texttt{stated} if the result is explicitly stated as a theorem, lemma, proposition, corollary, claim, fact or observation. Otherwise \texttt{intermediate}.\\
\texttt{statement\_lines} & For a stated result, where the parser finds its statement. For an intermediate conclusion, each sentence that asserts it, with its displays.\\
\texttt{proof\_lines} & The ranges from where the paper starts to establish the conclusion to where it is established, including deferred parts. Empty if the paper gives no proof.\\
\texttt{summary} & For a stated result, its statement as the parser finds it. For an intermediate conclusion, the conclusion in at most two sentences.\\
\texttt{evidence} & One quotation. For a stated result, we quote from its proof. For an intermediate conclusion, we quote the clause that asserts it.\\
\texttt{cited\_as} & \texttt{result} or \texttt{conjecture} if the statement, or the sentence that introduces it, credits an outside source. Otherwise empty.\\
\texttt{attribution\_content} & Empty if \texttt{cited\_as} is empty. Otherwise the quoted attribution, such as a citation in the header of a result.\\\bottomrule
\end{tabularx}
\end{table}

\begin{table}[htb]
\centering
\caption{The attributes of each edge.}
\label{tab:rdg-edge-schema}
\begin{tabularx}{\linewidth}{@{}p{.27\linewidth}X@{}}\toprule
Attribute & Content\\\midrule
\texttt{source} & The identifier of the result used as a direct premise.\\
\texttt{target} & The identifier of another result whose written argument directly uses the conclusion of the source.\\
\texttt{use\_lines} & The ranges where the target argument uses the premise.\\
\texttt{evidence} & One verbatim quotation where the target argument uses the premise.\\\bottomrule
\end{tabularx}
\end{table}

\subsection{Example}
\label{app:gips-example}
\label{app:rdg-worked-example}

The example in Section~\ref{subsec:GIPS} is the paper of \citet{WuXiaXu2022Floquet}, which we picked at random among the Human papers of about median length. Its Markdown has 84,147 bytes (55th percentile), and its GIPS has 77 nodes (49th percentile) and 77 edges (50th percentile). Figure~\ref{fig:app-real-rdg} shows its complete GIPS. On this GIPS, the example measurements of Table~\ref{tab:feature-families} work out as follows. The three cited results are Lemma~2.1, Lemma~2.2 and Theorem~4.4 ($3/77\approx0.039$). In 8 of the 77 edges, the statement of the premise first appears later in the text than the statement of its target ($8/77\approx0.104$). There are 64 intermediate conclusions and 13 stated results ($64/13\approx4.92$). Theorem~4.1 has the largest in-degree, 8, and the mean in-degree is 1 ($8/(77/77)=8$). The only feed-forward triangle is Corollary~2.2 $\to$ I40 $\to$ Theorem~4.1 with Corollary~2.2 $\to$ Theorem~4.1, among 95 directed two-step walks ($1/95\approx0.011$). Finally, the largest weakly connected component has 53 nodes ($53/77\approx0.688$). The other components have 6, 5 and 2 nodes, and 11 nodes are isolated. Table~\ref{tab:example-nodes} lists the attributes of the four nodes on the path in Figure~\ref{fig:workflow} from Corollary~2.2 through I44 and I45 to Theorem~4.1.

\begingroup
\setlength{\tabcolsep}{4pt}
\setlength{\abovedisplayskip}{0pt}
\setlength{\belowdisplayskip}{0pt}
\setlength{\abovedisplayshortskip}{0pt}
\setlength{\belowdisplayshortskip}{0pt}
\setlength{\LTpre}{6pt}
\setlength{\LTpost}{0pt}
\begin{longtable}{@{}>{\raggedright\arraybackslash}p{.115\linewidth}>{\raggedright\arraybackslash}p{.865\linewidth}@{}}
\caption{The four nodes of the path Corollary~2.2 $\to$ I44 $\to$ I45 $\to$ Theorem~4.1, with text as in the article.}\label{tab:example-nodes}\\
\toprule
Attribute & Content\\\midrule
\endfirsthead
\toprule
Attribute & Content\\\midrule
\endhead
\bottomrule
\endlastfoot
\texttt{id} & Corollary~2.2\\
\texttt{kind} & \texttt{stated}\\
\texttt{summary} & Corollary 2.2. Let $\mathbb{T}$ be an arbitrary time scale and $c$ be an arbitrary nonnegative constant. Then we have
\[\int_{a}^{b}\int_{a}^{t_{1}}\cdots\int_{a}^{t_{n-1}}c~\Delta t_{n}\cdots\Delta t_{1}\leq\frac{c(b-a)^{n}}{n!},\]
where $a,b\in\mathbb{T},a\leq t_{n-1}\leq\cdots\leq t_{1}\leq b.$\\
\texttt{evidence} & 2. Now suppose $k\leq n-1$ and that $S(k)$ holds. Then
{\[\begin{split}&\int_{a}^{t_{n-(k+1)}}\cdots\int_{a}^{t_{n-1}}c~\Delta t_{n}\cdots\Delta t_{n-k}\\
\leq&\int_{a}^{t_{n-(k+1)}}\frac{c(t_{n-k}-a)^{k}}{k!}\Delta t_{n-k}\leq\int_{a}^{t_{n-(k+1)}}\frac{c(t_{n-k}-a)^{k}}{k!}\mathrm{d}t_{n-k}=\frac{c(t_{n-(k+1)}-a)^{k+1}}{(k+1)!}.\end{split}\]}
Thus, $S(k+1)$ holds.\\\midrule
\texttt{id} & I44\\
\texttt{kind} & \texttt{intermediate}\\
\texttt{summary} & The vector sequence uniformly converges to $\left(\begin{matrix}x(t)\\y(t)\end{matrix}\right)$, and $\bar{y}_{n}(t)$ uniformly converges to $\bar{y}(t)$.\\
\texttt{evidence} & Finally we have $\left(\begin{matrix}x_{n}(t)\\y_{n}(t)\end{matrix}\right)$ uniformly converges to $\left(\begin{matrix}x(t)\\y(t)\end{matrix}\right)$ and $\bar{y}_{n}(t)$ uniformly converges to $\bar{y}(t)$.\\\midrule
\texttt{id} & I45\\
\texttt{kind} & \texttt{intermediate}\\
\texttt{summary} & The coefficient relation is
\[\mathcal{A}=\sum\limits_{n=0}^{\infty}\mathcal{A}_{n}.\tag{30}\]\\
\texttt{evidence} & By $\mathcal{A}=x(t_{0}+T)+\bar{y}(t_{0}+T)$ and Eq.~(24), we get
\[\mathcal{A}=\sum\limits_{n=0}^{\infty}\mathcal{A}_{n}.\tag{30}\]\\\midrule
\texttt{id} & Theorem~4.1\\
\texttt{kind} & \texttt{stated}\\
\texttt{summary} & Theorem 4.1. The expression of $\mathcal{A}$ mentioned in Theorem 3.1 is
{\[\begin{split}\mathcal{A}=&\displaystyle{\left(1+\dfrac{\phi(t_{0}+T)}{\phi(t_{0})}\right)\cos_{\phi}(t_{0}+T)}\\
&+\displaystyle{\int_{t_{0}}^{t_{0}+T}\left(\dfrac{1}{\phi(t_{0})}\cos_{\phi}(t_{1})Q(t_{0}+T,t_{1})-\sin_{\phi}(t_{1})P(t_{0}+T,t_{1})\right)\phi(t_{1})h(t_{1})\Delta t_{1}}\\
&+\displaystyle{\sum\limits_{n=2}^{\infty}\int_{t_{0}}^{t_{0}+T}\int_{t_{0}}^{t_{1}}\cdots\int_{t_{0}}^{t_{n-1}}\left(\dfrac{1}{\phi(t_{0})}\cos_{\phi}(t_{n})Q(t_{0}+T,t_{1})-\sin_{\phi}(t_{n})P(t_{0}+T,t_{1})\right)\cdot}\\
&\displaystyle{\phi(t_{n})Q(t_{n-1},t_{n})\cdots Q(t_{1},t_{2})\prod\limits_{i=1}^{n}h(t_{i})\Delta t_{n}\cdots\Delta t_{1}},\end{split}\]}\hfill(35)\newline
and the expression of $\mathcal{B}$ is
\[\mathcal{B}=e_{-p+\mu q}(t_{0}+T,t_{0}).\]\\
\texttt{evidence} & Thus we have
{\[\begin{split}\mathcal{A}=&\displaystyle{\left(1+\dfrac{\phi(t_{0}+T)}{\phi(t_{0})}\right)\cos_{\phi}(t_{0}+T)}\\
&+\displaystyle{\int_{t_{0}}^{t_{0}+T}\left(\dfrac{1}{\phi(t_{0})}\cos_{\phi}(t_{1})Q(t_{0}+T,t_{1})-\sin_{\phi}(t_{1})P(t_{0}+T,t_{1})\right)\phi(t_{1})h(t_{1})\Delta t_{1}}\\
&+\displaystyle{\sum\limits_{n=2}^{\infty}\int_{t_{0}}^{t_{0}+T}\int_{t_{0}}^{t_{1}}\cdots\int_{t_{0}}^{t_{n-1}}\left(\dfrac{1}{\phi(t_{0})}\cos_{\phi}(t_{n})Q(t_{0}+T,t_{1})-\sin_{\phi}(t_{n})P(t_{0}+T,t_{1})\right)\cdot}\\
&\displaystyle{\phi(t_{n})Q(t_{n-1},t_{n})\cdots Q(t_{1},t_{2})\prod\limits_{i=1}^{n}h(t_{i})\Delta t_{n}\cdots\Delta t_{1}}.\end{split}\]}\hfill(32)\\
\end{longtable}
\endgroup

\section{Landscape and signature analyses}
\label{app:controls}

In this appendix, we give the details of Sections~\ref{sec:landscape} and~\ref{sec:signatures}.
\subsection{Features and statistical tests}
\label{app:features-tests}

\subsubsection{Structural features}
\label{sec:feature_registry}

We compute each structural feature from the GIPS of one paper. A feature can use the nodes, the edges and the kind of each node. It can also use the labels of stated results, such as Lemma~2.4. From these labels we read result types, numbering and sections. In addition, a feature can use the attribution of nodes and the recorded locations of statements, proofs and uses. The GIPS records each location as a range of the 200-character lines of Appendix~\ref{sec:response_schema}. For the attribution of nodes, we call a node cited if its \texttt{cited\_as} attribute (Table~\ref{tab:rdg-node-schema}) is nonempty. If this attribute is \texttt{result}, we call the node a cited result. A stated cited result without a recorded proof span is called a reference node. For each of the six families, Table~\ref{tab:app-feature-families} lists the five features with the largest $\lvert d\rvert$ in the paired comparison of AI-rebuilt and Human papers (Appendix~\ref{app:paired-generation}).

\begingroup
\newcommand*{\featdef}[1]{\multicolumn{5}{@{\hspace{1em}}>{\raggedright\arraybackslash}p{\dimexpr\linewidth-1em\relax}@{}}{#1}}
\begin{longtable}{@{}>{\raggedright\arraybackslash}p{\dimexpr\linewidth-168pt-8\tabcolsep\relax}>{\raggedleft\arraybackslash}p{34pt}>{\raggedleft\arraybackslash}p{46pt}>{\raggedleft\arraybackslash}p{30pt}>{\raggedleft\arraybackslash}p{58pt}@{}}
\caption{The five structural features of each family with the largest $\lvert d\rvert$ in the paired AI-rebuilt--Human comparison. Human and AI-rebuilt are the cohort means, $d$ is the standardized mean difference, and $q$ is the Benjamini--Hochberg adjusted $p$-value of the paired test.}
\label{tab:app-feature-families}\\
\toprule
Feature & Human & AI-rebuilt & $d$ & $q$ \\
\midrule
\endfirsthead
\toprule
Feature & Human & AI-rebuilt & $d$ & $q$ \\
\midrule
\endhead
\bottomrule
\endfoot
\multicolumn{5}{@{}l}{\emph{References}}\\*
Externally cited-result fraction & 0.061 & 0.002 & $-1.39$ & $4.1\times10^{-103}$ \\*
\featdef{Fraction of nodes that are externally cited results.}\\
Cited-node fraction & 0.061 & 0.002 & $-1.38$ & $1.4\times10^{-102}$ \\*
\featdef{Fraction of nodes that are cited.}\\
Cited-to-uncited edge fraction & 0.066 & 0.002 & $-1.28$ & $3.2\times10^{-92}$ \\*
\featdef{Fraction of edges whose premise is cited and whose target is uncited.}\\
Cited intermediate-conclusion fraction & 0.044 & 0.002 & $-1.13$ & $2.1\times10^{-75}$ \\*
\featdef{Fraction of intermediate conclusions that are cited.}\\
Uncited-to-uncited edge fraction & 0.899 & 0.994 & 1.12 & $1.2\times10^{-75}$ \\*
\featdef{Fraction of edges joining two uncited nodes.}\\
\addlinespace
\multicolumn{5}{@{}l}{\emph{Organization}}\\*
Length of shorter statements & 3.99 & 3.21 & $-2.02$ & $3.7\times10^{-181}$ \\*
\featdef{25th percentile of statement lengths among nodes with a recorded statement location, measured in units of 100 normalized characters.}\\
Multiple nodes sharing the same statement location & 0.018 & 0.084 & 1.96 & $2.9\times10^{-173}$ \\*
\featdef{Sum nodes beyond the first at each nonempty location. Divide by the total number of nodes.}\\
Proof length per premise & 2.26 & 1.66 & $-1.65$ & $1.1\times10^{-132}$ \\*
\featdef{Mean of $\log(1+\ell/k)$ over the stated results with a recorded proof span and $k\ge1$ direct premises, where $\ell$ is the proof length in units of 100 normalized characters.}\\
Mean log proof length & 2.15 & 1.85 & $-1.65$ & $2.9\times10^{-142}$ \\*
\featdef{Mean of $\log(1+\ell)$ over the nodes with a recorded proof span, where $\ell$ is the proof length.}\\
Length of longer proofs & 2.92 & 2.36 & $-1.54$ & $7.4\times10^{-126}$ \\*
\featdef{90th percentile of $\log(1+\ell)$ over the nodes with a recorded proof span, where $\ell$ is the proof length.}\\
\addlinespace
\multicolumn{5}{@{}l}{\emph{Node Statistics}}\\*
Stated results without recorded proof & 0.211 & 0.002 & $-1.41$ & $1.5\times10^{-106}$ \\*
\featdef{Fraction of stated results with no recorded proof span.}\\
Numbering gaps & 0.130 & 0.008 & $-1.16$ & $1.3\times10^{-51}$ \\*
\featdef{Fraction of second numbers skipped between the smallest and the largest among the stated-result labels with the same leading number, averaged over the leading numbers with at least two distinct second numbers, for example $1/4$ for Lemma~2.1, Theorem~2.3 and Corollary~2.4.}\\
Intermediate-conclusion fraction & 0.841 & 0.904 & 0.88 & $5.1\times10^{-52}$ \\*
\featdef{Fraction of nodes that are intermediate conclusions.}\\
Stated-result fraction & 0.159 & 0.096 & $-0.88$ & $5.1\times10^{-52}$ \\*
\featdef{Fraction of nodes that are stated results.}\\
Corollaries without recorded proof & 0.270 & 0.024 & $-0.88$ & $7.2\times10^{-18}$ \\*
\featdef{Fraction of corollaries with no recorded proof span.}\\
\addlinespace
\multicolumn{5}{@{}l}{\emph{Edge Statistics}}\\*
Stated results with intermediate premises & 0.583 & 0.872 & 1.34 & $2.1\times10^{-109}$ \\*
\featdef{Fraction of stated results with a direct intermediate premise.}\\
Premises per proved result & 2.58 & 3.57 & 0.94 & $2.8\times10^{-59}$ \\*
\featdef{Mean number of direct premises of the stated results with a recorded proof span.}\\
Premises per corollary & 1.50 & 2.71 & 0.89 & $1.9\times10^{-18}$ \\*
\featdef{Mean number of direct premises of corollaries.}\\
Fraction of nodes that are terminal stated results & 0.038 & 0.016 & $-0.86$ & $4.2\times10^{-48}$ \\*
\featdef{Number of stated results with at least one premise that are not used by any other node, divided by the total number of nodes.}\\
Unused stated results & 0.323 & 0.170 & $-0.82$ & $8.3\times10^{-47}$ \\*
\featdef{Fraction of stated results that are not used by any other node.}\\
\addlinespace
\multicolumn{5}{@{}l}{\emph{Graph Motifs}}\\*
Transitive-to-cyclic triad ratio & 8.05 & 15.14 & 0.78 & $1.5\times10^{-47}$ \\*
\featdef{(Transitive triangles $+\,1$) / (cyclic triangles $+\,1$).}\\
Connected triples per node & 2.42 & 3.35 & 0.63 & $1.8\times10^{-41}$ \\*
\featdef{Number of node triples connected after ignoring edge directions, divided by the number of nodes.}\\
Normalized two-step walks & 1.07 & 1.21 & 0.57 & $7.4\times10^{-26}$ \\*
\featdef{Directed two-step walks $u\to v\to w$, divided by $m^2/n$ for a GIPS with $n$ nodes and $m$ edges.}\\
Feed-forward triangles per edge & 0.067 & 0.095 & 0.51 & $1.3\times10^{-22}$ \\*
\featdef{Feed-forward triangles, formed by $u\to v\to w$ and $u\to w$, divided by the number of edges.}\\
Bi-fan-to-fork ratio & 0.178 & 0.092 & $-0.49$ & $1.1\times10^{-18}$ \\*
\featdef{(Bi-fans $+\,1$) / (forks $+\,1$). A bi-fan is two nodes at most eleven places apart in the order in which the GIPS lists its nodes, together with two results that use both nodes as direct premises. A fork is a node together with two results that use it as a direct premise.}\\
\addlinespace
\multicolumn{5}{@{}l}{\emph{Global Topology}}\\*
Mean incoming harmonic reachability & 2.21 & 3.17 & 0.84 & $1.5\times10^{-72}$ \\*
\featdef{Mean of sums of $1/d$ from nodes reaching each node, where $d$ is the shortest directed path length.}\\
Katz centrality Gini coefficient & 0.059 & 0.070 & 0.80 & $4.2\times10^{-56}$ \\*
\featdef{Gini coefficient of the Katz centralities $z_v$ of the nodes, where $z_v$ starts at 1 and is updated six times to $1+0.1\sum_u z_u$ over the direct premises $u$ of $v$.}\\
Mean Katz centrality & 1.11 & 1.13 & 0.77 & $1.3\times10^{-61}$ \\*
\featdef{Mean of $z_v$ over the nodes.}\\
Number of weak components per node & 0.242 & 0.143 & $-0.76$ & $6.4\times10^{-54}$ \\*
\featdef{Number of connected components after ignoring edge directions, divided by the total number of nodes.}\\
Largest-to-mean Katz centrality & 1.54 & 1.79 & 0.74 & $1.0\times10^{-40}$ \\*
\featdef{Largest $z_v$ divided by the mean $z_v$.}\\
\end{longtable}
\endgroup

\subsubsection{Effect sizes and tests}
\label{app:statistical-specifications}

Each comparison has two cohorts, for example AI-rebuilt papers and Human papers, which we call cohort~1 and cohort~0. For each feature, we only use the papers in which the feature is defined. As the effect size, we use the signed standardized mean difference
\[
d=\frac{m_1-m_0}{\sqrt{(s_1^2+s_0^2)/2}},
\]
where $m_g$ is the mean and $s_g^2$ is the sample variance in cohort~$g$.
\begin{itemize}[leftmargin=*,itemsep=3pt]
\item \textbf{Unpaired comparisons.} We test the mean difference with the Welch $t$-test.
\item \textbf{Paired comparisons.} The paired $t$-test uses the within-pair differences $D_i=x_{1,i}-x_{0,i}$ over the $N$ complete pairs and compares $t=\bar D/(s_D/\sqrt{N})$ with a $t$ distribution with $N-1$ degrees of freedom. If all differences are zero, we set $t=0$ and $p=1$. For $d$, we compute $m_g$ and $s_g^2$ over the complete pairs.
\end{itemize}
To compare the marginal distributions of a feature, we use the two-sided two-sample Kolmogorov--Smirnov test, over the complete pairs in a paired comparison. In each comparison, we apply the Benjamini--Hochberg procedure separately to the structural and the linguistic features, and separately to the tests of means and the Kolmogorov--Smirnov tests. We call a difference \emph{significant} when its adjusted $p$-value $q<0.05$.

\subsection{Structural landscape}
\label{app:landscape-details}

\subsubsection{Principal components and relative positions}

On PC1, PC2 and PC3 of Figure~\ref{fig:exp4pf-landscape}a,b, the cohort means are $(-1.65,-0.73,-1.22)$ for Human, $(-2.09,-0.97,-0.66)$ for AI-reworded, $(-0.36,0.33,0.32)$ for Hybrid, $(3.17,1.89,1.78)$ for AI-rebuilt and $(7.44,0.73,2.74)$ for AI papers. PC2 and PC3 explain 8.1\% and 5.7\% of the variance.

For Figure~\ref{fig:exp4pf-landscape}c, we summarize $r$ over the 285 features. AI-reworded papers remain close to Human papers, with an interquartile range of $r$ from $-0.17$ to $0.16$. They move away from AI-rebuilt on 155 features and extend beyond AI-rebuilt on 4, and their median by family runs from $0.21$ for Organization down to $-0.12$ for Graph Motifs. Hybrid papers occupy an intermediate position, with an interquartile range from $0.02$ to $0.41$. They move in the AI-rebuilt direction on 220 features and extend beyond AI-rebuilt on 15, and their median by family runs from $0.35$ for References down to $0.18$ for Node Statistics. AI papers depart much further from Human papers, with an interquartile range from $0.31$ to $1.54$. Their median by family is $0.31$ for Organization, $0.82$ for References and between $1.01$ and $1.41$ for the other four families.

\subsubsection{Length-matched landscape and temporal positions}
\label{app:matched-landscape}
\label{app:temporal-variation}

To build the length-matched sample, we first collected as candidates all AI-rebuilt and AI papers, the Human and Hybrid papers of the training set with distinct texts, and the AI-reworded papers. Second, we split the manuscript lengths into bands of 500 words. In each band, we take the smallest number of candidates in any cohort and call it the capacity of the band. When we count the Human candidates, we leave out the Human sources of AI-rebuilt papers. The capacities sum to the same number of papers in each cohort, all between 1,500 and 7,000 words. Finally, we drew the cohorts in the order AI-reworded, AI-rebuilt, AI, Human and Hybrid, using a fixed-seed PCG64 permutation within each band. When we drew the Human papers, we skipped those whose AI-rebuilt version had already been drawn.

\begin{figure}[!ht]
\centering
\includegraphics[width=.85\linewidth,keepaspectratio,trim=0 11.5bp 0 0,clip]{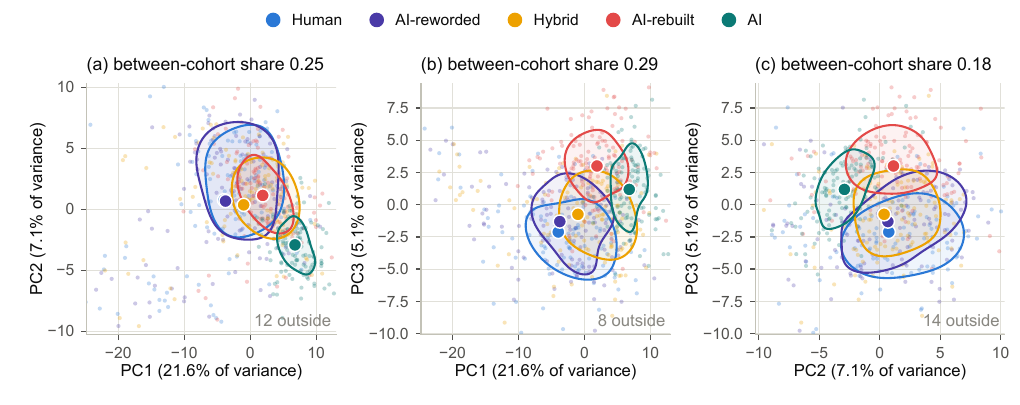}
\caption{PC1--PC2, PC1--PC3 and PC2--PC3 projections of the papers in the length-matched landscape. Faint points are papers, contours enclose approximately the densest half of each cohort, and larger dots show cohort means. The panel titles give the between-cohort share $\sum_c n_c\lVert\bar z_c-\bar z\rVert^2/\sum_i\lVert z_i-\bar z\rVert^2$, where $z_i$ is the point of paper $i$, $\bar z$ the mean of all points, and $n_c$ and $\bar z_c$ the size and mean of cohort $c$. The number marked ``outside'' in each panel counts the papers beyond its displayed range.}
\label{fig:app-matched-pca}

\vspace{10pt}
\includegraphics[width=\linewidth,height=.36\textheight,keepaspectratio]{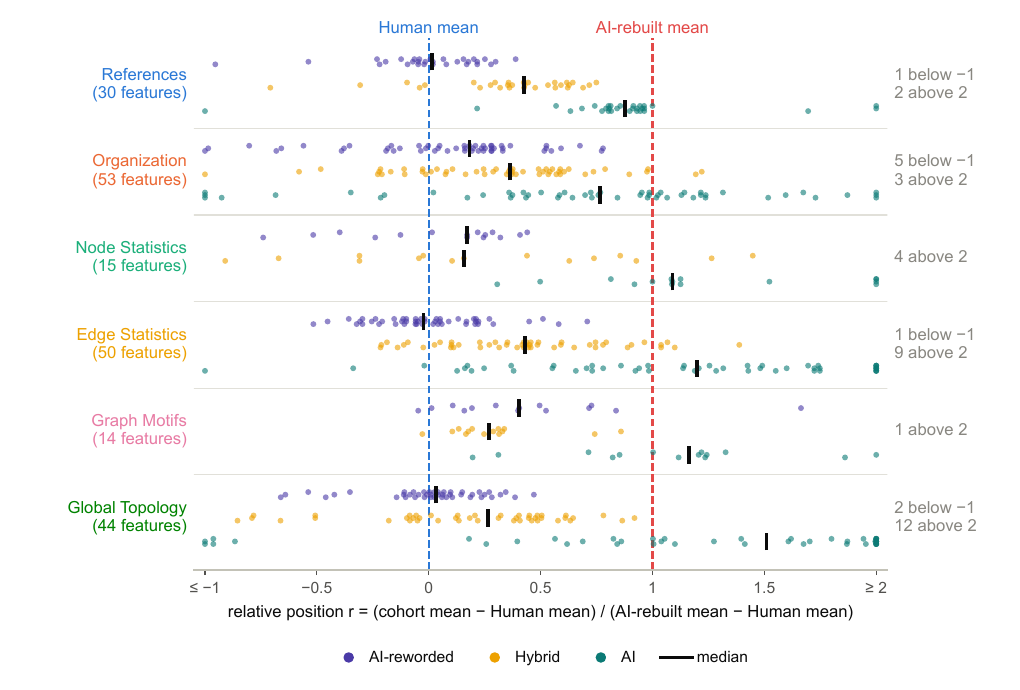}
\caption{Relative positions of cohort means in the length-matched landscape. Dots show individual features and black bars show medians across features, by family (counts in parentheses).}
\label{fig:app-matched-axis}
\end{figure}

\begin{figure}[!ht]
\centering
\includegraphics[width=\linewidth,keepaspectratio]{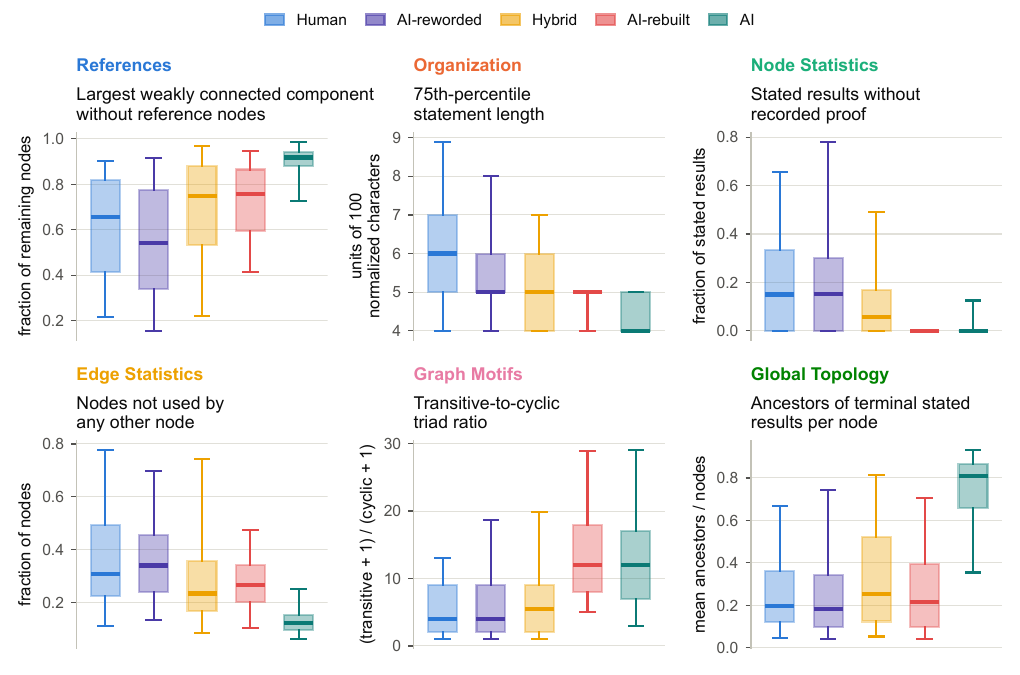}
\caption{Six structural features in the five cohorts of the length-matched landscape, one from each family. Boxes show interquartile ranges, inner lines show medians, and whiskers reach the 5th and 95th percentiles.}
\label{fig:app-matched-features}
\end{figure}

The PCA of the matched papers (Figure~\ref{fig:app-matched-pca}) follows the full-data PCA above, except that we standardize the features with the mean and standard deviation of the matched Human papers. For the relative positions (Figure~\ref{fig:app-matched-axis}), we test the structural features in the matched sample and keep the 206 with a significant unpaired AI-rebuilt--Human difference. The medians of $r$ are 0.05 for AI-reworded, 0.36 for Hybrid and 0.99 for AI papers. For Figure~\ref{fig:app-matched-features}, we pick in each family the feature with the largest Welch $F$ statistic, which compares the means of the five cohorts.

For the temporal positions of Hybrid papers in Figure~\ref{fig:disclosed-temporal-positions}, we put each Hybrid paper in the period of its first version and record three covariates. The first is the research field, which has eight levels: the six fields of Section~\ref{subsec:controlled-comparisons}, other mathematics, and other fields. The second is the reported AI-use group. We go through four groups in this order and put the paper in the first one that applies according to the paper: AI contributed mathematics (ideas, problems, definitions, proofs or examples), AI checked the mathematics, AI formalized it, or AI was itself the research method. The third is the revised-version status, which records whether the version we read is later than the first one.

\begin{figure}[!ht]
\centering
\includegraphics[width=\linewidth,keepaspectratio]{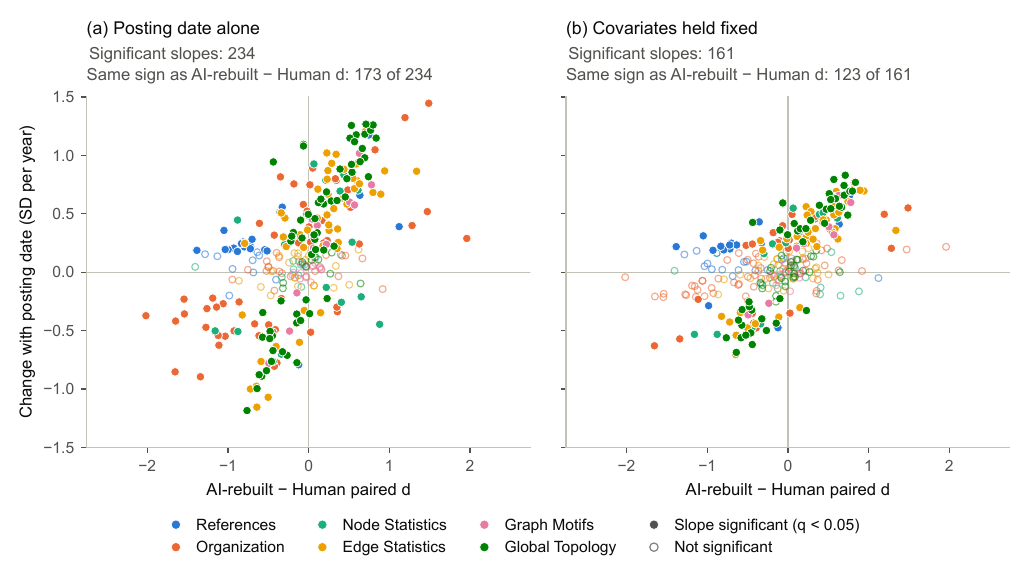}
\caption{Change of each structural feature with posting date in Hybrid papers, against its paired AI-rebuilt--Human $d$. Panel (a) uses posting date alone, and panel (b) holds research field, reported AI-use group and revised-version status fixed. Each dot is one feature, colored by family. Filled dots are significant slopes, and a dot in the upper right or lower left quadrant changes in the direction of $d$.}
\label{fig:app-temporal-slopes}
\end{figure}

We also test each structural feature and each of the 37 linguistic features for a change with posting date. For each feature, we fit a straight line of the feature on the date of the first version, in years, over the Hybrid papers that have a value. We fit it once with this date alone and once with the three covariates held fixed. Dividing the slope by the feature's standard deviation gives a change in standard deviations per year. We test each slope with HC2 standard errors and call it significant when its Benjamini--Hochberg $q$ is below 0.05, within the structural and within the linguistic features. With the covariates held fixed, 161 structural slopes are significant, and 123 of them have the sign of the paired AI-rebuilt--Human $d$ (Figure~\ref{fig:app-temporal-slopes}). For the linguistic features, the two counts are 20 and 13.

\subsection{Recurring signatures}
\label{app:signature-details}

\subsubsection{Paired AI-rebuilt--Human comparison}
\label{app:paired-generation}

In Figure~\ref{fig:exp1}a, we plot the unadjusted paired-test $p$-value of each feature at height $\log(1-\log_{10}p)$. With this scale, $p$-values that span more than 180 orders of magnitude fit in one panel. Table~\ref{tab:app-feature-families} gives the definitions of the twelve features of Figure~\ref{fig:signature-interpretations}. Two of them, in panels c and d, depend on how the GIPS records statement locations (Appendix~\ref{sec:feature_registry}). As a result, statement lengths take only a few distinct values, and several extracted conclusions in one passage share the same recorded statement location.

\subsubsection{Recurrence and robustness}
\label{app:recurrence}
\label{app:field-adjustment}
\label{app:length-matching}
\label{app:field-agreement}

\begin{figure}[!ht]
\centering
\includegraphics[width=.93\linewidth]{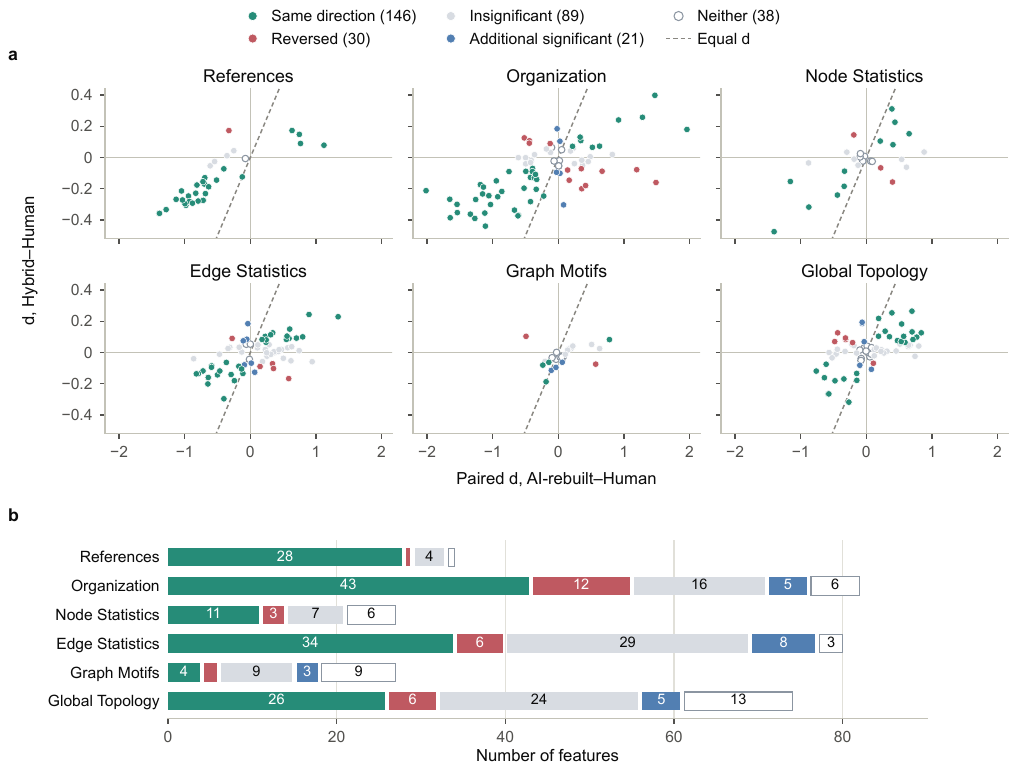}
\caption{Classes of the structural features in the Hybrid--Human comparison. (a) Paired AI-rebuilt--Human $d$ against Hybrid--Human $d$, by family. Green, red and gray features are in the reference set. Their Hybrid--Human difference is significant with the same sign (green), significant with the opposite sign (red) or not significant (gray). Blue features are outside the reference set and have a significant difference. Open circles are features significant in neither comparison. (b) Number of features in each class, by family.}
\label{fig:app-hybrid-recurrence}
\end{figure}

\begin{figure}[!ht]
\centering
\includegraphics[width=.93\linewidth]{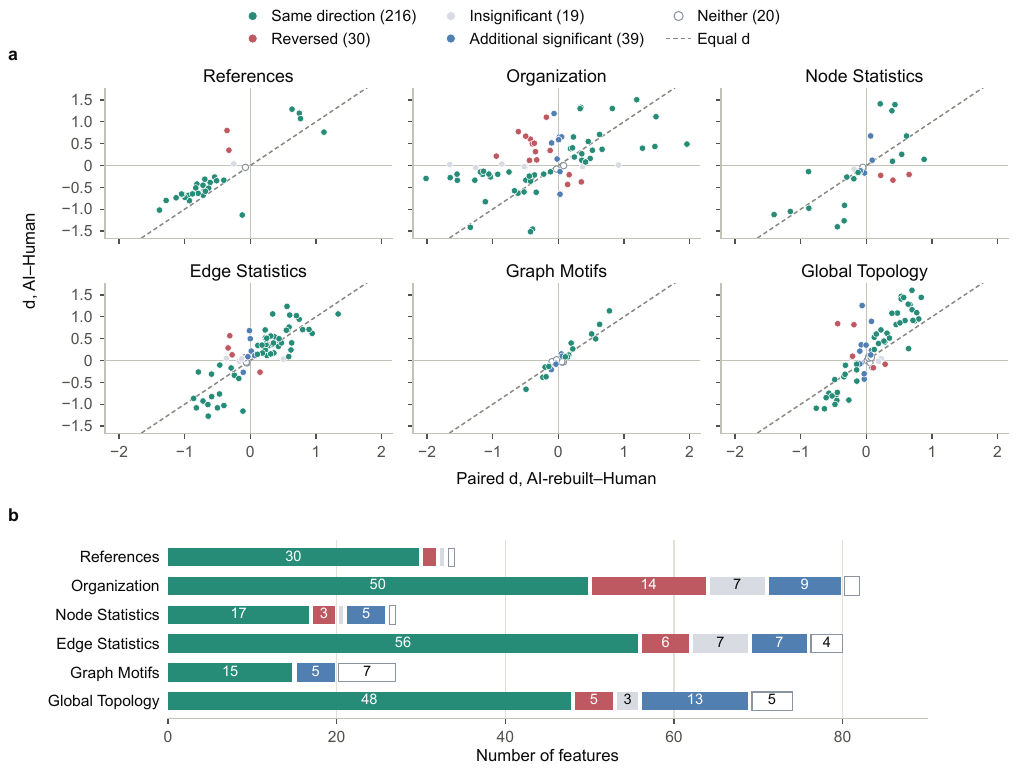}
\caption{Classes of the structural features in the AI--Human comparison, drawn as in Figure~\ref{fig:app-hybrid-recurrence}.}
\label{fig:app-ai-recurrence}
\end{figure}

For Figure~\ref{fig:persistence}a, we compare the 2,489 Hybrid papers with the 2,491 Human papers of the training set by Welch $t$-tests. We compare the 722 AI papers with the same Human papers in the same way. These Human papers include the 700 sources of AI-rebuilt papers. In each comparison, the structural features form one Benjamini--Hochberg family. We read each of these comparisons against a reference comparison, here the paired AI-rebuilt--Human comparison. Its 265 significant features form the reference set. We then place each feature in one of the five classes defined in the caption of Figure~\ref{fig:app-hybrid-recurrence}. Figure~\ref{fig:app-hybrid-recurrence} and Figure~\ref{fig:app-ai-recurrence} show these classes.

The field and length controls compare Hybrid and Human papers in the six-field analysis. This analysis uses the papers of the training set in the six research fields of Section~\ref{subsec:controlled-comparisons}. We test the structural features by Welch $t$-tests. The 166 significant features form the reference set of Figure~\ref{fig:persistence}b. Under each field or length control, we place these features in the five classes above. For the field adjustment, we estimate one Hybrid--Human difference within fields and assume that it is the same in all six fields. Its $d$ uses the same denominator as the unadjusted $d$.

\begin{figure}[!ht]
\centering
\includegraphics[width=.74\linewidth]{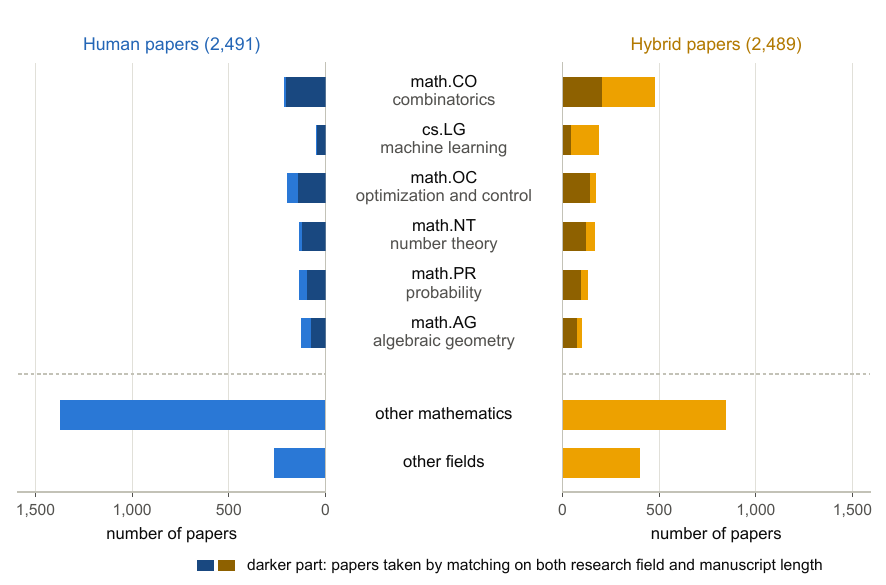}
\caption{Human and Hybrid papers of the training set by research field. The darker part of each bar counts the papers taken by matching on both research field and manuscript length.}
\label{fig:app-field-composition}
\end{figure}

For the matching controls, we make two matched draws, each with the same number of papers per cohort, and compare the two cohorts of each draw by Welch $t$-tests. We put a paper with a manuscript length of $w$ words in the 500-word band $\lfloor w/500\rfloor$. In matching on both research field and manuscript length, we take from each of the 152 cells (one field and one band) as many papers of each cohort as the smaller cohort has there (Figure~\ref{fig:app-field-composition}). In length-only matching, we take in each of the 41 bands as many papers of each cohort as the first draw has there, but from all six fields together. In both draws, each cell or band gets the first of its papers in a fixed-seed PCG64 permutation of the cohort.

Within each of the six fields, we repeat the Hybrid--Human comparison, with Benjamini--Hochberg adjustment within that field. We also repeat the paired AI-rebuilt--Human comparison in the five fields with at least 20 pairs, using the field of each pair's Human paper. To check agreement with the AI-rebuilt profile, we ask whether the Hybrid--Human and AI-rebuilt--Human differences agree in direction, before and after field adjustment. A feature agrees in direction when its two differences have the same sign, even if neither is significant. To keep the Human reference sets of the two contrasts disjoint, the AI-rebuilt--Human contrast uses the pairs whose Human paper is in one of the six fields. The Hybrid--Human contrast uses the Hybrid papers and the Human papers in these fields that are not sources of AI-rebuilt papers.

\label{app:graph-size}The graph-size adjustment uses the 700 Human--AI-rebuilt pairs. In pair $i$, let $D_i$, $\Delta\log(1+n_i)$ and $\Delta\log(1+m_i)$ be the AI-rebuilt minus Human differences in a feature, in $\log(1+n)$ and in $\log(1+m)$, where $n$ and $m$ are the numbers of nodes and edges of a GIPS. For each feature, we regress $D_i$ on the two size differences, with an intercept, over its $N$ complete pairs. The intercept is the difference between an AI-rebuilt paper and a Human paper whose GIPSs have the same numbers of nodes and edges. We divide it by the denominator of the unadjusted $d$ to get the adjusted $d$, and we test it by a $t$-test with HC2 standard errors (i.e., each pair may have its own error variance) and $N-3$ degrees of freedom. We place the features in the same classes, with the 265 significant features of the paired comparison as the reference set. The adjusted-to-unadjusted effect ratio of a feature is its adjusted $d$ divided by its unadjusted $d$.

\section{Detection experiment}
\label{app:detection}
\label{app:release-benchmark}

For ProofTrace, we feed the 342 features into a network with hidden layers of 128 and 64 GELU units and dropout 0.2. We train the network on the 4,980 papers of the training set for four epochs with seed 1, using AdamW (learning rate $10^{-3}$, weight decay $10^{-4}$, batches of 128, cosine schedule). ProofTrace++ computes its 200 structural and linguistic features from the GIPS and the full Markdown text of a paper. We fill missing values and rescale the features with values stored with the model, and we add 60 indicators of missing values. We feed them into three networks, each with hidden layers of 128 and 64 ReLU units, batch normalization and dropout 0.5. The three networks use seeds 1, 2 and 3 and were trained for 16, 15 and 8 epochs, respectively. The score is the mean of their AI-oriented log-odds. The pretrained and zero-shot baselines work as follows.
\begin{itemize}[leftmargin=*,itemsep=3pt]
\item \textbf{MELD (pretrained)} uses the released detector \citep{reword_meld_card}, which scores each token of a text in a style space against learned anchors of human writing and prototypes of AI model families. For a chunk of text, the released head takes the mean of the highest quarter of the token scores. To score a paper, we cut it down to its main body and split this into balanced, nonoverlapping chunks of at most 2,046 text tokens. The paper score is the mean of the chunk scores, weighted by their numbers of text tokens.
\item \textbf{ModernBERT-MAGE (pretrained)} reads the whole paper in windows of up to 8,190 text tokens, and we add the model's start and end tokens to each window. For each window, a softmax with the released temperature turns the two class scores of the released head into the probability that the text is AI-generated. The paper score is the mean of these probabilities, weighted by the number of text tokens in each window.
\item \textbf{DETree (pretrained)} uses the released RealBench reference bank. From each paper, we take at most eight prose passages of at most 512 tokens each. For each passage, we get an embedding by maximum pooling over the tokens of encoder layer 19 and scale it to unit length. We then score the passage against its $k=5$ nearest examples in the bank. To get the paper score, we keep the highest-scoring quarter of the passages and aggregate their scores.
\item \textbf{ToS (pretrained)} uses the released Longformer-Motif detector, which classifies scenes of a text with motif features from the DMRST parser. We apply it to every scene of the complete paper, using 414 features that record the frequency and depth of motifs. The paper score is the unweighted mean of the machine-class probabilities of the scenes.
\item \textbf{RAIDAR (zero-shot)} uses GPT-5.6 Luna to rewrite each complete paper. The score is $1-d/\ell$, where $d$ is the character-level Levenshtein distance between the two texts and $\ell$ is the length of the longer one.
\end{itemize}

The fitted and adapted baselines use the training set of 2,491 Human and 2,489 Hybrid papers. For each detector, we select the settings by fitting on part of these papers and validating on the rest. We then refit the detector with these settings on all 4,980 papers.
\begin{itemize}[leftmargin=*,itemsep=3pt]
\item \textbf{ModernBERT-MAGE (LR)} cuts the paper into nonoverlapping windows of 512 tokens, including the model's start and end tokens. In each window, we mean-pool the token embeddings of the frozen encoder, and we average the window vectors with equal weights. We standardize these vectors and set the regularization of the logistic regression to $C=0.1$, selected on validation. The score is the fitted log-odds of the Hybrid class.
\item \textbf{ModernBERT-MAGE (finetuned)} uses the same document vectors. Its head starts from the released values and is trained with AdamW (learning rate $10^{-4}$, weight decay 0.01, batches of 128 papers) for 50 epochs, selected on validation. The score is the difference between the head's Hybrid and Human outputs.
\item \textbf{RADAR (finetuned)} starts from the released detector \texttt{TrustSafeAI/RADAR-Vicuna-7B}, a RoBERTa-large encoder with a classifier head. We combine the frozen encoder's outputs over 512-token windows into one vector per paper and train the head on them for 29 epochs, selected on validation.
\item \textbf{Desklib (finetuned)} starts from the released detector \texttt{desklib/ai-text-detector-v1.01} \citep{reword_desklib_card}. We build the paper vectors as for RADAR (finetuned) and train the head with AdamW for 50 epochs, selected on validation.
\item \textbf{DETree (kNN)} replaces the released reference bank with the embeddings and labels of the 4,980 papers of the training set, and validation selects layer 18 and $k=50$. To embed a paper, we take its first 510 text tokens plus the model's start and end tokens, apply maximum pooling over the tokens of this layer and scale the result to unit length. We then let the $k$ training papers nearest to it by inner product vote, each weighted by a softmax of its similarity. The score is the weighted share of Hybrid papers among them.
\end{itemize}

\end{document}